\RequirePackage{fix-cm}
\documentclass[twocolumn]{svjour3}          
\smartqed  
\usepackage{color}
\usepackage{subcaption}
\usepackage{amsmath}
\usepackage{colortbl}
\usepackage{caption}
\usepackage{multicol}
\usepackage{graphicx}
\usepackage{multirow}%
\usepackage{amsmath,amssymb,amsfonts}%
\usepackage{mathrsfs}%
\usepackage[title]{appendix}%
\usepackage{textcomp}%
\usepackage{manyfoot}%
\usepackage{booktabs}%
\usepackage{algorithm}%
\usepackage{algorithmicx}%
\usepackage{algpseudocode}%
\usepackage{listings}%
\usepackage{float}

\usepackage{natbib}
\usepackage[dvipsnames]{xcolor}
\definecolor{citecolor}{HTML}{0071bc}
\usepackage[colorlinks,citecolor=citecolor]{hyperref}
\newcommand{\rotbox}[1]{\rotatebox{60}{#1}}
\begin{document}

\title{Progressive Visual Prompt Learning with Contrastive Feature Re-formation
}

\author{Chen Xu \and Yuhan Zhu
\and Haocheng Shen  \and Boheng Chen
\and Yixuan Liao \and Xiaoxin Chen
\and Limin Wang }


\institute{Chen Xu \at
              State Key Laboratory for Novel Software Technology, Nanjing University,  Nanjing, China \\
              \email{chenxu24568@gmail.com}           
           \and
           Yuhan Zhu \at
              State Key Laboratory for Novel Software Technology, Nanjing University,  Nanjing, China \\
              \email{zyuhan0812@gmail.com}
           \and 
           Haocheng Shen \at
              Vivo AI Lab, Shenzhen, China \\
              \email{haocheng.shen@vivo.com}
           \and
           Boheng Chen \at
              Vivo AI Lab,  Shenzhen, China \\
              \email{eechenboheng@gmail.com}
           \and
           Yixuan Liao \at
              Vivo AI Lab,  Shenzhen, China \\
              \email{yixuan.liao@vivo.com}
           \and
           Xiaoxin Chen \at
              Vivo AI Lab,  Shenzhen, China \\
              \email{xiaoxin.chen@vivo.com}
           \and
           Limin Wang \at
            State Key Laboratory for Novel Software Technology, Nanjing University, Nanjing, China \\
            Shanghai AI Laboratory, Shanghai, China \\
              \email{lmwang@nju.edu.cn}
           \and
}

\date{Received: date / Accepted: date}

\maketitle

\begin{abstract}
Prompt learning has recently emerged as a compelling alternative to the traditional fine-tuning par-adigm for adapting pre-trained Vision-Language (V-L) models to downstream tasks.  Drawing inspiration from the successful application of prompt learning in Natural Language Processing, pioneer research efforts have been predominantly concentrated on text-based prompting strategies. By contrast, the potential of visual prompting within V-L models remains underexploited. The straightforward transposition of existing visual prompt methods, tailored for Vision Transformers (ViT), into the V-L models often leads to suboptimal performance or training instability, highlighting the unique challenges of this research avenue. 
To mitigate these challenges, in this paper, we propose a novel structure called \textbf{Pro}gressive \textbf{V}isual \textbf{P}rompt (\textbf{ProVP}). This design aims to strengthen the interaction among prompts from adjacent layers, thereby enabling more effective propagation of image embeddings to deeper layers in a manner akin to an instance-specific manner. Additionally, to address the common issue of generalization deterioration in the training period of learnable prompts, we further introduce a contrastive feature re-formation technique for visual prompt learning. This method prevents significant deviations of prompted visual features from the fixed CLIP visual feature distribution, ensuring better generalization capability.
Combining the \textbf{ProVP} and the contrastive feature \textbf{Re-f}ormation technique, our proposed method, \textbf{ProVP-Ref}, significantly stabilizes the training process and enhances both the adaptation and generalization capabilities of visual prompt learning in V-L models post-training. To demonstrate the efficacy of our approach, we evaluate ProVP-Ref across 11 image datasets, achieving the state-of-the-art results on \textbf{7} of these datasets in both few-shot learning and base-to-new generalization settings. 
To the best of our knowledge, this is the first study to showcase the exceptional performance of visual prompts in V-L models compared to previous text prompting methods in this area. Code will be available at \textbf{\url{https://github.com/MCG-NJU/ProVP}}.
\end{abstract}

\section{Introduction}
\label{sec:intro}
Vision-Language (V-L) pre-trained models, such as CLIP \citep{CLIP}, have demonstrated remarkable potential across a wide range of tasks~\citep{gao2021clip,zhang2021tip,conde2021clip,li2021language,wang2022cris,ghiasi2022scaling,gu2021open,du2022learning,shi2022proposalclip}, such as image recognition, semantic segmentation, objection detection, etc. Unlike traditional models that rely on closed-set category labels, V-L models are trained to align visual and textual features with a web-scale corpus of image-text pairs.  Leveraging the power of natural language guidance, these models are able to learn open-set visual concepts and exhibit strong generalization capabilities.

Inspired by the success of prompt learning in Natural Language Processing~\citep{P-tuning,li2021prefix}, prompt learning techniques~\citep{CoOp,CoCoOp,ProGrad} has recently been employed to adapt V-L models to downstream tasks by utilizing a small set of learnable prompts, while keeping the pre-trained model fixed. Existing research has mainly focused on text-based approaches such as CoOp~\citep{CoOp}, CoCoOp~\citep{CoCoOp}, ProGrad~\citep{ProGrad}. However, incorporating prompts into the visual side of V-L models received less attention. It is expected that integrating prompts into the visual pathway can enhance the ability of pre-trained V-L models to handle shifts in the image data distribution and adapt more effectively to new domains.

\begin{figure}[tbp]
    \centering
    \includegraphics[width=0.48\textwidth]{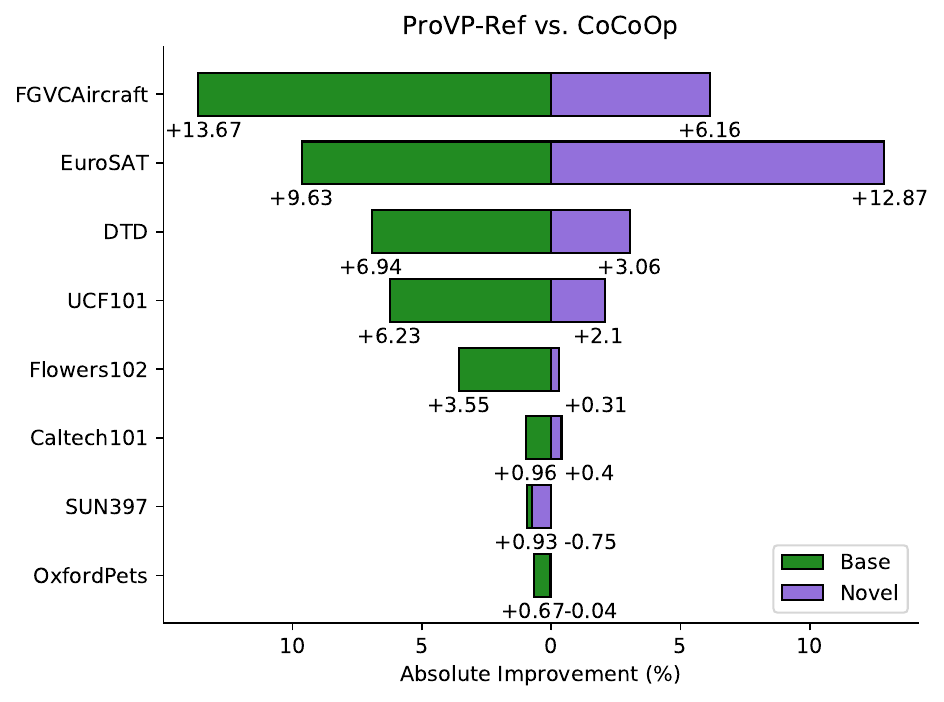}
    \caption{ Absolute gains of ProVP-Ref over CoCoOp~\citep{CoCoOp} in base-to-new generalization on 8 datasets. ProVP-Ref demonstrates a significant improvement in both base and new categories.}
    \label{fig.B2NDiff}
\end{figure}

\begin{figure}[tbp]
    \centering
    \includegraphics[width=0.4\textwidth]{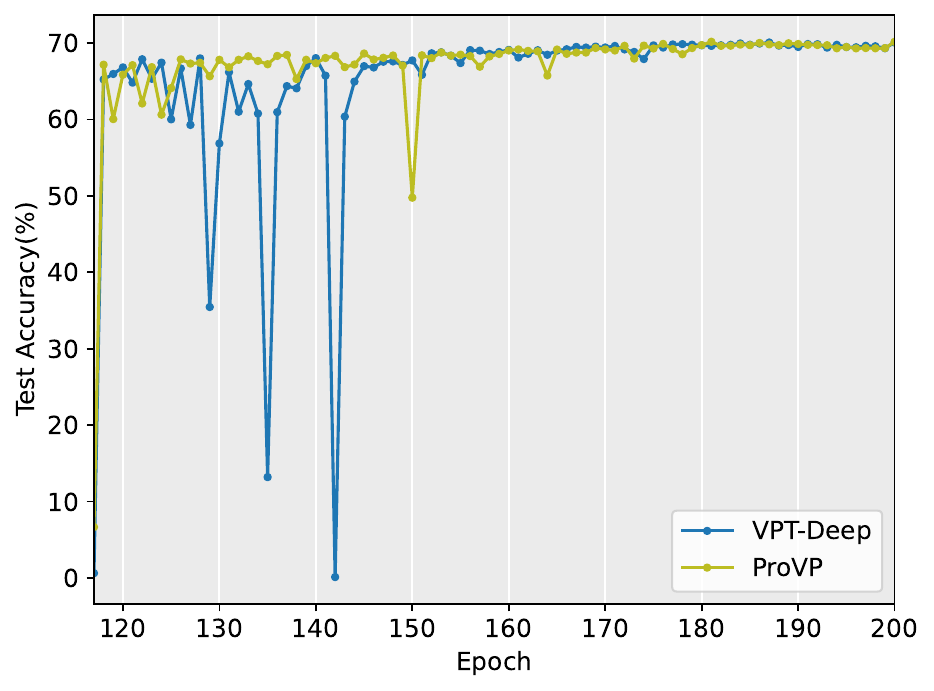}
    \caption{Test accuracy curve of VPT-Deep and ProVP during training (we select the results of 8-shot learning on ImageNet). The Deep version of VPT shows serious training instability as the tested performance could be dropped several times even close to 0.}
    \label{fig.train}
\end{figure}

From the limited literature~\citep{bahng2022exploring,VPT}, optimizing visual prompts for V-L models remains challenging with current designs. For example,~\citep{bahng2022exploring} implemented visual prompts by padding learnable pixels but failed to yield downstream performance improvements comparable to full fine-tuning. VPT~\citep{VPT} introduced a deep visual prompt tuning (VPT) approach for single-modal Vision Transformer (ViT)~\citep{vit} models. The training phase for Visual Prompt Tuning (VPT) variants has been observed to suffer from instability and sensitivity to hyperparameters. Specifically, throughout the training process, the test performance exhibits significant volatility (As illustrated in Fig.\ref{fig.train}, the tested performance of VPT experiences severe declines at multiple points during training). We also observed that the outcomes of training VPTs are disproportionately affected by the hyperparameters such as learning rate or the length of prompts, as even minor adjustments to these hyperparameters can lead to a substantial drop in tested performance. Taken together, these issues make training VPT variants prone to frequent crashes, thereby complicating the search for optimal tuned performance. In this paper, we find that a contributing factor to this instability may be the isolated learning of prompts within each layer of the Vision Transformer (ViT), which are initialized randomly and fail to foster necessary interactions during learning. This isolation can induce significant disturbances in the pre-trained model's functioning, amplifying the likelihood of overfitting. In addition, despite MaPLe 's~\citep {MaPLe} advancement in proposing a multi-modal prompt learning approach that incorporates visual modality, these persistent challenges have not been addressed. This oversight suggests a critical gap in our understanding of the training and adaptation for visual prompts in V-L models, signaling the need for more sophisticated, interaction-conscious prompt learning frameworks.

\begin{figure*}[tbp]
  \centering
  \includegraphics[width=1\textwidth]{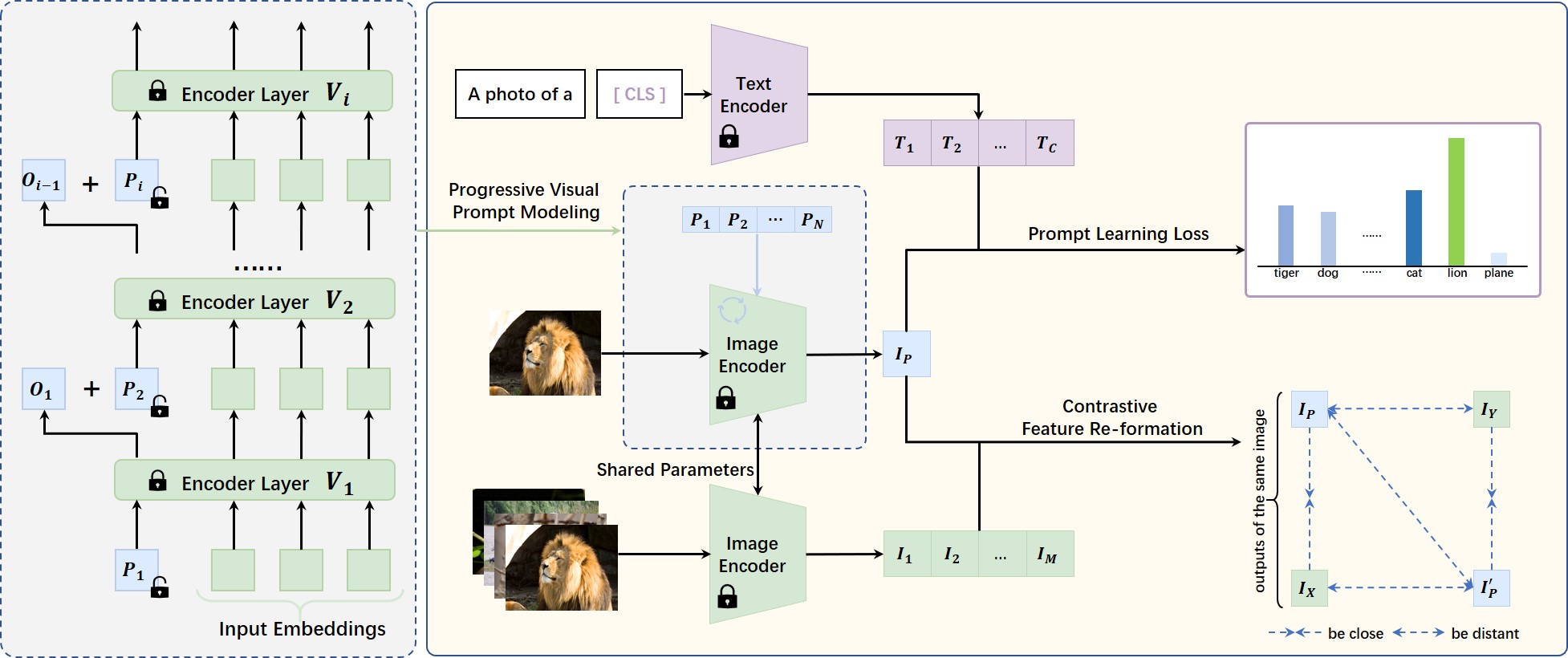}
  \caption{An overview of ProVP-Ref. Right: a full pipeline for our approach. It consists of two main parts Progressive Visual Prompt, in which we retain the outputs of the prompt via a progressive connection, and Contrastive Feature Re-formation: Using the frozen image encoder and the original image
  input, we can reformate the prompted features so that they can constitute a more similar representation with pre-trained CLIP. Left: a detailed view of Progressive Visual Prompt, new prompts will be combined with the output of the former embeddings before being sent to the encoder layer.}
  \label{fig:overview}
\end{figure*}

Building on the aforementioned analysis, this paper introduces a novel structure called \textbf{Pro}gressive \textbf{V}isual \textbf{P}rompt (\textbf{ProVP}). Figure~\ref{fig:overview} shows the overview of our approach. ProVP establishes connections between prompts across adjacent layers. Specifically, the prompt embedding at each layer is a combination of the newly inserted prompts and the output of the prompt embedding from the previous layer.
This ProVP structure enhances the interaction between prompts from different layers, leading to improved effectiveness, efficiency, and stability of prompt learning. Furthermore, ProVP facilitates the propagation of image embeddings to deeper layers and exhibits characteristics akin to instance-specific prompting through progressive connections between adjacent prompts. This mitigates the risk of overfitting and enhances the model's robustness to domain shifts.

In addition to the ProVP structure, we address the issue of generalization deterioration in prompt learning, where the model's ability to recognize unseen classes significantly deteriorates after training on downstream tasks. To mitigate this problem, we propose a novel contrastive feature re-formation method incorporating constraints on the prompted visual features. Given that randomly initialized prompts can cause a notable shift in pre-trained features, we maintain the model's generalization capability by re-formating the prompted output features according to the visual feature distribution in CLIP. 
Different from existing works~\citep{ProGrad} and~\citep{LASP} which leverage the prediction logits of zero-shot CLIP, our method preserves the pre-trained knowledge in the \textit{feature space}, which is verified to be more effective for visual modality and offers more flexibility in handling downstream tasks with substantial domain gaps. Combining both, ProVP-Ref shows a consistent improvement in both adaptation and generalization ability, demonstrating the effectiveness of our methods. 

The main contributions are summarized as follows:
\begin{itemize}
\item We investigate the utilization of visual prompts with CLIP and propose a novel Progressive Visual Prompt (ProVP) structure. To the best of our knowledge, we are the first to demonstrate the effectiveness of single-modal visual prompts in V-L pre-trained models, exhibiting superior performance over previous prompt-based methods in downstream tasks (Figure~\ref{fig.B2NDiff}). 
	
\item To alleviate the generalization deterioration problem in visual prompt learning, we propose a new contrastive feature re-formation method that prevents the significant deviation of prompted visual features from the CLIP visual feature distribution. With carefully designed feature constraints, our mo-del improves adaptation and generalization capabilities simultaneously.

\item We evaluate our method on 11 image benchmark datasets and achieve 7 out of 11 state-of-the-art results in both few-shot and base-to-new settings. Notably, our method provides significant gains over text prompt counterparts on the datasets that have larger distribution shifts.
\end{itemize}

\section{Related Work}\label{relatedwork}

\subsection{Vision-Language Models}
Vision-Language (V-L) models~\citep{CLIP, jia2021scaling, yuan2021florence, zhai2022lit, li2021supervision} have gained significant popularity in the computer vision community due to their robust representation and remarkable generalization capabilities.
The pre-training of these models involves various pretext tasks, including BERT-like masked-language modeling~\cite{Vilbert, Vilt}, masked-region modeling~\cite{Vl-bert, Lxmert}, and contrastive learning~\citep{CLIP, jia2021scaling, zhai2022lit, EVA-CLIP}.
CLIP~\citep{CLIP} and ALIGN~\citep{jia2021scaling} are two representative methods in this domain, which consist of a visual encoder and a text encoder and are pre-trained to align visual and text-encoded features via contrastive learning. Thanks to vast amounts of pre-training text-image corpora (e.g., approximately 400M for CLIP and around 1B for ALIGN), they have shown great potential in various downstream tasks, such as image classification \citep{gao2021clip,zhang2021tip,conde2021clip}, semantic segmentation \citep{li2021language,wang2022cris,ghiasi2022scaling}, object detection \citep{gu2021open,du2022learning,shi2022proposalclip} and image captioning \citep{mokady2021clipcap,tang2021clip4caption,shen2021much}. In this work, we utilize CLIP as the foundation model and specifically focus on the challenging task of image classification in the few-shot learning scenario.

\subsection{Prompt Learning}
Prompt learning methods were first proposed in natural language processing~\citep{P-tuning,li2021prefix} to replace the handcraft engineering for text templates, which introduce learnable token embeddings optimized through the model training as additional inputs for the model. These learned prompts are utilized to help generate more contextually relevant outputs for language models and have shown remarkable performance in related fields. Recently, prompt learning has also been extended into the visual modality by Visual Prompt Tuning \citep{VPT}. VPT inserts learnable embeddings into each transformer layer of a ViT \citep{vit} model to help transfer the model to downstream tasks. For multi-modal modalities, most pioneering work of prompt learning for V-L models is inspired by the popularity of prompt learning in NLP and focuses on text encoder \citep{CoOp, CoCoOp,ProGrad,lu2022prompt,shu2022test}, while keeping image encoder untouched. 

\subsection{Prompt Learning in Vision-Language Models} 
Compared to prompt learning approaches in a single language or visual modality, adopting such a method in multi-modal scenarios is more challenging. CoOp \citep{CoOp} firstly proposes a learnable text prompt instead of the hand-crafted templates (e.g., 'A photo of a') to adapt the CLIP model for few-shot classification, and CoCoOp \citep{CoCoOp} then identifies the generalization problem of CoOp, as the performance of the tuned model highly drops on unseen categories. This \textbf{generalization deterioration} problem might be caused by the overfitting of such prompts and impairs the model's open-set classification capability. To handle this problem, CoCoOp developed an extra Meta-Net to produce instance-level prompts conditioned on each image. It was evaluated on a base-to-new setting to demonstrate its superior performance in unseen classes.
ProGrad \citep{ProGrad} only updates the prompt whose gradient is aligned to the hand-crafted prompt on zero-shot settings to prevent forgetting
the pre-trained knowledge of CLIP. 
ProDA \citep{lu2022prompt} extends CoOp by optimizing multiple text prompts simultaneously and estimates the distribution of the output 
embeddings of the prompts. 
TPT \citep{shu2022test} proposes test-time prompt tuning that learns adaptive text prompts on the fly with a single test sample. 
Compared with text prompt learning, visual prompt learning in V-L models is much less explored. 
Bahng \citep{bahng2022exploring} firstly explored visual prompts with CLIP in \textit{pixel space} by padding learnable pixels around the input images. 
Such pixel-level prompt is difficult to learn and the tuned performance is unsatisfactory compared to full fine-tuning.  
Recently, VPT \citep{VPT} was proposed to insert visual prompts in \textit{embedding space} into a ViT \citep{vit} in a shallow or deep manner where the prompts were inserted into each layer of ViT and learned independently. It was built on a single-modal model while the more powerful V-L models, such as CLIP, have not been attempted. 
Some other works \citep{wang2022learning,wang2022s} also make use of visual prompts but were invented specifically for incremental learning. MaPLe~\citep{MaPLe} and DPL~\citep{xu2023dpl} introduce a multi-modal prompt learning method but the difficulty of visual prompting is still less explored. Our work differs from these works as we focus on visual prompt learning with V-L models while the text encoder is kept frozen. A novel progressive visual prompt structure is also proposed which is different from VPT and has not been investigated before.  

\subsection{Knowledge Transfer} 
Although forgetting mitigation has been widely explored in incremental learning \citep{MCIncre,DistillIncre,regularRL,icarl,LearnWoForget,MTMI}, such methods can not directly be employed for prompt learning. In incremental learning, such works aim at continually learning new knowledge without forgetting the previous ones, avoiding the performance drop of previous tasks. In prompt learning, on the contrary, forgetting mitigation helps the model capture the relevant knowledge from the large-scale pre-trained knowledge of V-L models, which eventually benefits the performance on downstream tasks. Identifying such differences, ProGrad~\citep{ProGrad} uses a gradient align method to learn prompts without conflicting with the pre-trained knowledge (e.g., zero-shot CLIP predictions). However, over-reservation of general knowledge in prompt learning may distract the training process, which causes insufficient learning. We relieve this problem by constraining the learned prompts on the feature space via contrastive learning, which is more flexible for keeping useful information from pre-trained knowledge in V-L models.

\section{Methodology}
\label{sec:method} 
\subsection{Revisiting CLIP}
\noindent\textbf{CLIP}~\citep{CLIP} is comprised of a decoupled text and image 
encoder pair. The image encoder aims to encode the image to a low-dimensional feature representation and can be either a CNN-like model as ResNet-50~\citep{resnet}, or a ViT~\citep{vit}. The text encoder employs the Transformer encoder architecture designed in~\citep{Attention} to convert the original input text into a hidden textual representation.
CLIP is trained on a large-scale image-text dataset by using the language-image contrastive learning strategy. Specifically, given a mini-batch of text-image pairs, the model maximizes the similarity of the features from paired text and image while minimizing the unpaired ones. Supported by a large-scale and high-quality training set of 400 million image-text pairs, CLIP has obtained great generalization capability to handle challenging open-vocabulary problems such as zero-shot image recognition. When predicting the label of an image, CLIP generates the textual representation for the class labels by feeding them with hand-crafted templates (e.g., ``\texttt{a photo of a [cls]}'') into the text encoder. These sentences are utilized to obtain specific embeddings for each category, which are then leveraged to compute the classification scores as the cosine similarity with the image feature representations. Denote $\{y_i\}_C$ as the label set containing $C$ classes of a downstream task and $\{t_i\}_C$ as the corresponding text templates, and let $g$ and $f$ be the text and image encoder of CLIP, the recognition score of CLIP is calculated as below:
\begin{equation}
    \label{eq.CLIPinfer}
    \begin{aligned}
        p(y_i|x)=\frac{\exp(<f(x),g(t_i)>/\tau)}{\sum\limits^C_{j=1}\exp(<f(x),g(t_j)>)/\tau},
    \end{aligned}
\end{equation}
where $\tau$ is a temperature parameter learned by CLIP and $<\cdot,\cdot>$ denotes cosine similarity.

\subsection{Visual Prompt Tuning}
\noindent\textbf{Visual Prompting Tuning}~\citep{VPT} inserts learnable token 
embeddings (visual prompts) into the input latent space of ViT~\citep{vit} and tunes them while freezing model backbone. VPT proposes two types of visual prompting: VPT-Shallow and VPT-Deep, with the latter proving to perform better on transfer learning tasks.
Formally, denoting a collection of $m$ learnable $d$-dimension prompts as $P_{i}=\{p_i^k\in \mathbb{R}^d|k=1,2,...,m\}$ and the original input embeddings $X_{i-1}$ for the $i$-th layer $L_i$, VPT-Deep displays the prompt insertion in a ViT~\citep{vit} with $N$ 
layers as:
\begin{equation}
  \label{eq.VPT-Deep}
  [X_{i},\underline{\hbox to 3mm{}}] = L_i([X_{i-1},P_{i}]),\quad i=1,2,\ldots,N.
\end{equation}

Despite VPT being verified to be effective in visual backbones as ViT~\citep{vit}, we found the training of VPT-Deep in VLMs is much more challenging:
the independent learning of prompts in each layer may confuse the direction of optimization, increasing the difficulty of training and making the model sensitive to hyper-parameters. Additionally, the randomly initialized prompts at each layer in VPT are likely to cause significant perturbations to the model output, which increases the risk of catastrophic forgetting and overfitting. 

\subsection{Progressive Visuals Prompt Learning}
Observing from Eq.\ref{eq.VPT-Deep} in VPT-Deep, we identify one potential cause for the training instability: the inconsistency in the prompting strategy. As each prompt in VPT-Deep solely contributes to its respective layer's propagation, post the $i_{th}$ layer, the output for the inserted prompt $P_i$, denoted as $O_i$, is discarded.  The newly inserted prompts show no correlation with the outputs and are initialized randomly, which introduces significant disturbance to the prompted features. Conversely, these discarded prompt outputs may harbor abundant information about input instances and pre-trained knowledge and can be beneficial for subsequent computations. However, directly maintaining such outputs with newly inserted prompts for the next layers will cause the increasing length for learnable prompts in deeper layers, which can severely disturb the learned feature representation when the length becomes too large, and lead to misalignment of adjacent layers in feature space.  In practice, we observed that such a simple structure is challenging to converge and further compromises generalization capability. To more effectively retrieve these discarded prompts, we propose a novel prompt learning structure, \textbf{P}rogressive \textbf{V}isual \textbf{P}rompt (\textbf{ProVP}), which utilizes a progressive connection to combine newly inserted prompt embeddings and the former outputs. Formally, in ProVP, the prompting strategy of the visual encoder is reformulated as:
\begin{equation}
\begin{aligned}
&[X_1,O_1]=L_1([X_0,P_1]),\\
\end{aligned}
\end{equation}
for the first layer $L_1$, and 
\begin{equation}
\begin{aligned}
&\left[X_{i},O_{i}\right]=L_i([X_{i-1},(1-\alpha)P_i+\alpha O_{i-1}]), \ i \neq 1,
\end{aligned}
\end{equation}
for the next layers, where $\alpha$ is the progressive decay. Without consistently increasing the prompt length, our method could maintain the learned information in the former layers by combining the prompt outputs and the new prompts with the control of $\alpha$. Additionally, the locations of prompts were verified equivalent in~\citep{VPT} (e.g., $[X_i, P_{i+1}]$ works the same as $[P_{i+1}, X_i]$) and we followed the settings in~\citep{VPT} in to insert visual prompts between image embeddings and the `[\texttt{CLS}]' token. 

Compared to the learning strategy of VPT-Deep described in Eq.~\ref{eq.VPT-Deep}, ProVP acts in an instance-specific manner: the prompt output of the former block $O_i$ is deeply correlated to the image embeddings and varies with inputs while the new insert prompt $P_i$ is still input invariant. Thus, the prompt input for the next layer $L_i$ as $P'_i = (1 - \alpha)P_i + \alpha * O_i$ also varies across different images. Similar to CoCoOp, we find that this architecture prioritizes the instance-level information rather than focuses on a subset of classes, which helps to prevent overfitting and increase robustness to domain shifts. Benefitting from the mentioned advantages, ProVP has achieved better capability both in adaptation and generalization tasks compared to VPT variants. Furthermore, the progressive connection utilized in ProVP has strengthened the interaction between prompts of adjacent layers in the model, reducing the oscillation of performance and sensitivity to hyper-parameters, leading to a much more stable training  (More discussion can be found in further study).

During ProVP tuning, let $f_{\mathbf{p}}$ be the prompted image encoder and  $g$ be the text encoder. We optimize the model by minimizing the negative log-likelihood: 
\begin{equation}
  \label{eq.CELoss}
  \centering
  \begin{split}
  \mathrm{Loss}_{ce}(x) &= -\sum\limits^C_{i=1}y_i\log p(t_i|x),\\
  p(y_i|x) &= \frac{\exp(<f_{\mathbf{p}}(x),g(t_i)>/\tau)}{\sum\limits_{j=1}^C \exp(<f_{\mathbf{p}}(x),g(t_j)>/\tau)}, 
  \end{split}
\end{equation}
where $y$ denotes the one-hot ground-truth annotation and $\tau$ is the temperature parameter learned by CLIP.

\subsection{Contrastive Feature Re-formation}
When adapting pre-trained models to downstream tasks through prompt learning, the risk of \textit{generalization deterioration} is a prevalent concern, exemplified by the notable performance drop observed in models like CoOp when testing on unseen classes after training compared to zero-shot CLIP. One possible reason, as identified by~\citep{ProGrad}, is the improper learning of prompts. Relying \textit{solely} on the cross-entropy loss (Eq.~\ref{eq.CELoss}) during the learning process may lead the model to forget pre-trained general knowledge and sub-optimally focus on specific downstream data, consequently compromising the inherited generalization ability from the pre-trained model. Motivated by the insights of \citep{ProGrad} and knowledge distillation methods ~\citep{Knowledge_Distillation1, Knowledge_Distillation2}, we address this forgetting problem by leveraging the pre-trained information from zero-shot CLIP and specifically design a new training strategy, Contrastive Feature Re-formation, for visual prompt learning. Different from preserving the zero-shot CLIP predictions as utilized in~\citep{ProGrad}, we maintain the model's generalizability under the guidance of the pre-trained image feature distribution on \textit{feature space}. As the randomly initialized prompts in visual modality cause a significant shift in the pre-trained features, the diversity of the prompted features may be reduced after training and be less distinguishable to the text encoder. Realizing that overcoming this problem could be more effective in alleviating the generalization deterioration, we introduce a new training strategy as Contrastive Feature Re-formation to re-formate the shifted features to constitute a similar distribution to pre-trained CLIP. Let $f$, $f_{\mathbf{p}}$ be the pre-trained and prompted image encoder, and $\{x_i\}^M_{i=1}$ represents a mini-batch of $M$ images. Contrastive feature re-formation constrains the prompted and pre-trained features of the same image $x_i$ to be close and the different ones to be distant. Therefore, the reformatting loss as $Loss_{Ref}$ is defined as
\begin{equation}
  \label{eq.ReformatLoss}
  \mathrm{Loss}_{Ref}(x_i) = -\log\frac{\exp(<f_{\mathbf{p}}(x_i),f(x_i)>)}{\sum\limits_{j=1}^M \exp(<f_{\mathbf{p}}(x_i),f(x_j)>)}.
\end{equation}
 Combining with the conventional cross entropy loss in Eq.~\ref{eq.CELoss}, the total training loss can be formulated as :
\begin{equation}
  \label{eq.Loss}
  \centering
  \mathrm{Loss}=\mathrm{Loss}_{ce}+\lambda \mathrm{Loss}_{Ref},
\end{equation}
where $\lambda\in [0,1]$ is a hyper-parameter to adjust the weight of the $Loss_{Ref}$ during training.

\begin{figure*}[tbp]
  \centering
  \begin{subfigure}[htbp]{0.32\linewidth}
    \includegraphics[width=\linewidth]{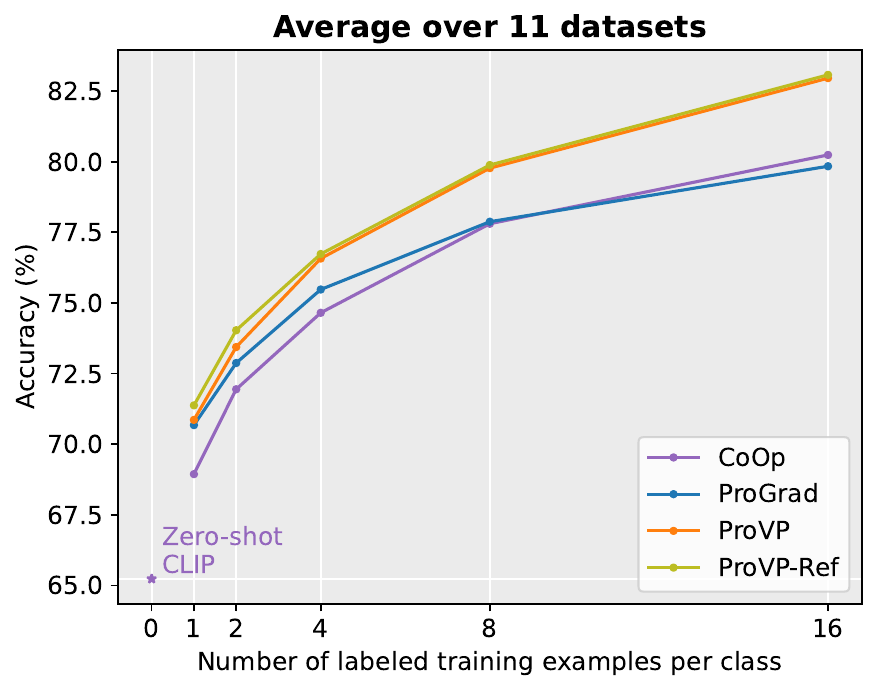}
  \end{subfigure}
  \hfill
  \begin{subfigure}[htbp]{0.32\linewidth}
    \includegraphics[width=\linewidth]{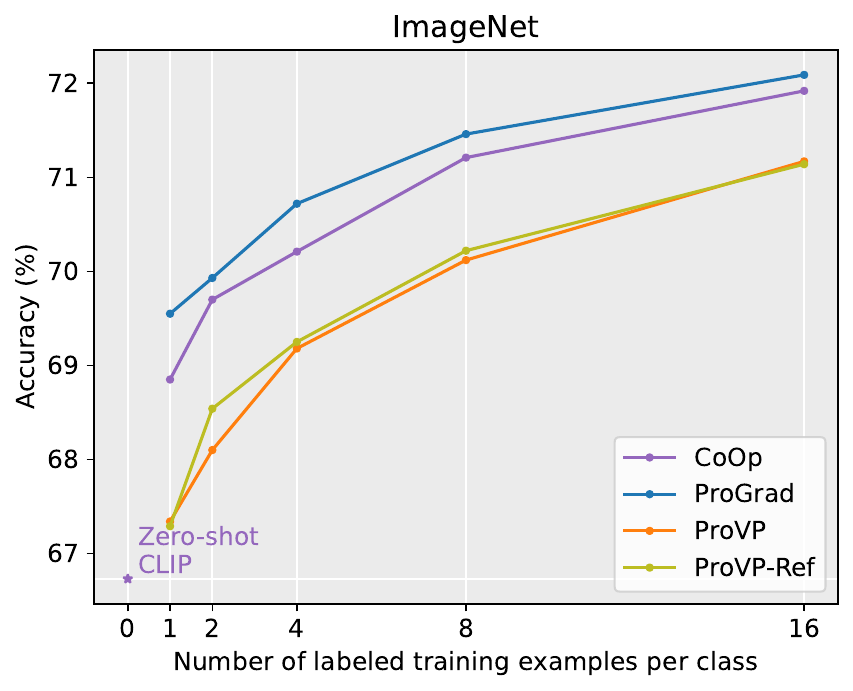}
  \end{subfigure}
  \hfill
  \begin{subfigure}[htbp]{0.32\linewidth}
    \includegraphics[width=\linewidth]{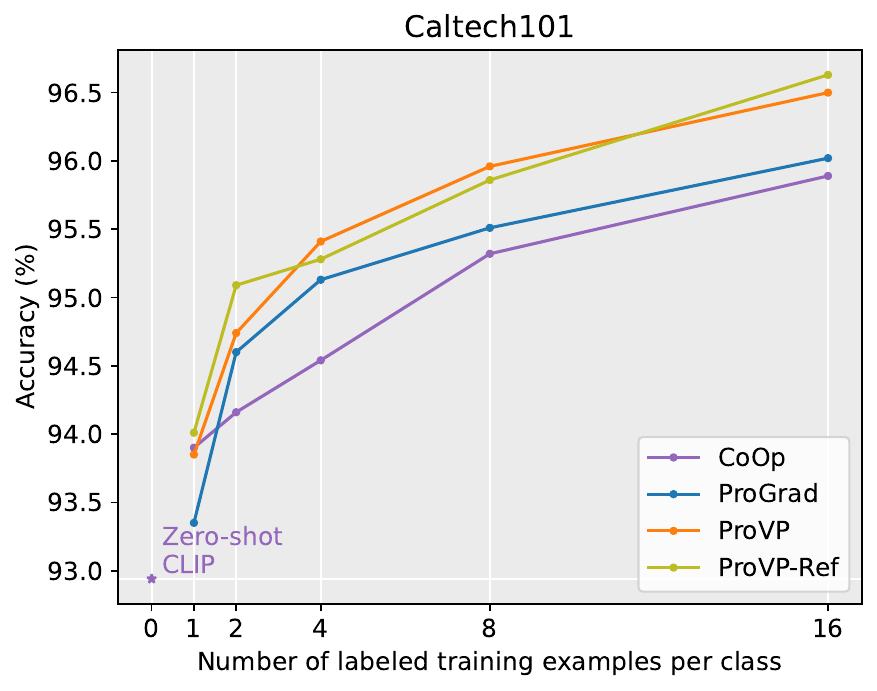}
  \end{subfigure}
  
  \begin{subfigure}[htbp]{0.32\linewidth}
    \includegraphics[width=\linewidth]{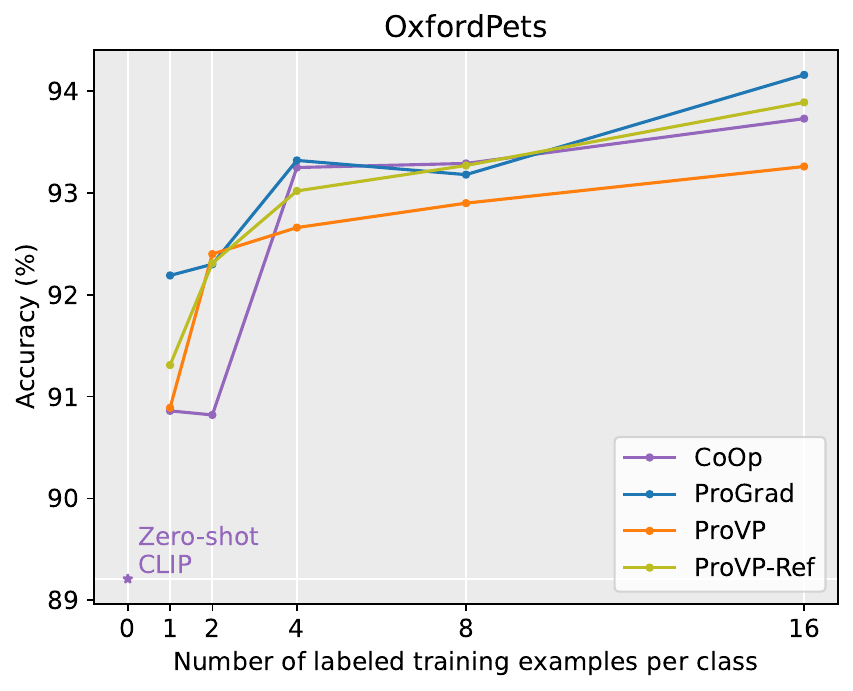}
  \end{subfigure}
  \hfill
  \begin{subfigure}[htbp]{0.32\linewidth}
    \includegraphics[width=\linewidth]{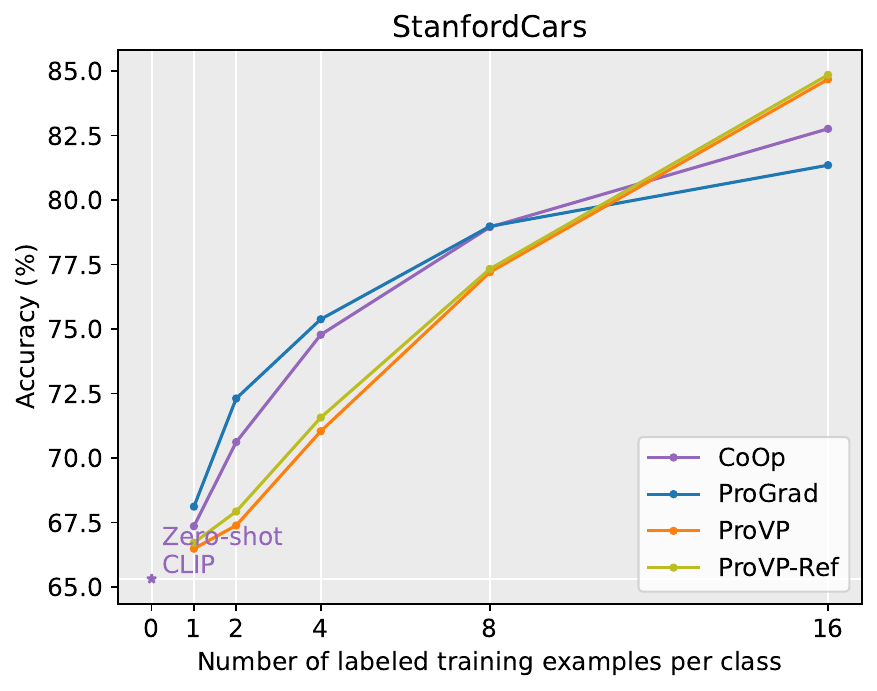}
  \end{subfigure}
  \hfill
  \begin{subfigure}[htbp]{0.32\linewidth}
    \includegraphics[width=\linewidth]{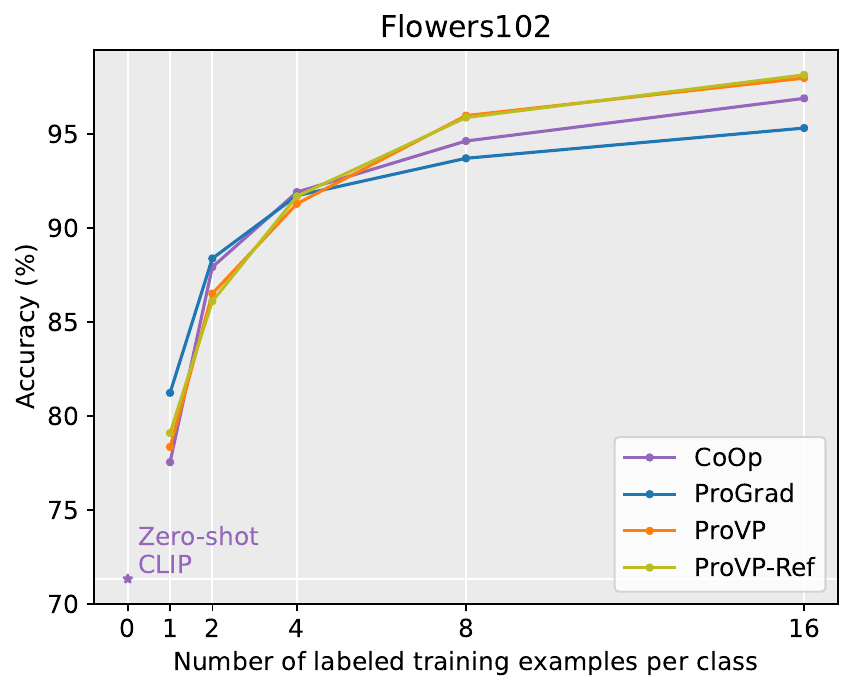}
  \end{subfigure}
  
  \begin{subfigure}[htbp]{0.32\linewidth}
    \includegraphics[width=\linewidth]{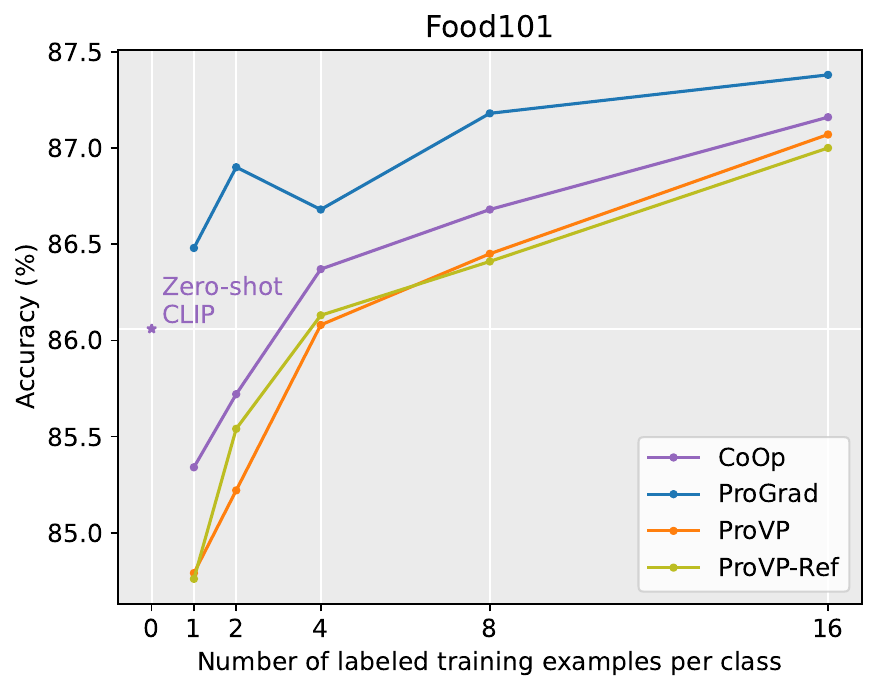}
  \end{subfigure}
  \hfill
  \begin{subfigure}[htbp]{0.32\linewidth}
    \includegraphics[width=\linewidth]{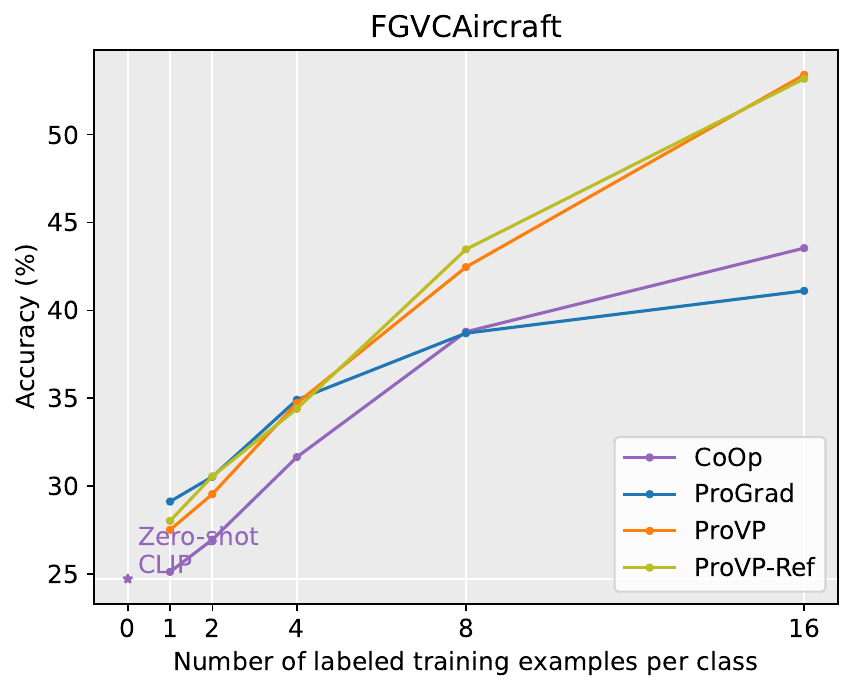}
  \end{subfigure}
  \hfill
  \begin{subfigure}[htbp]{0.32\linewidth}
    \includegraphics[width=\linewidth]{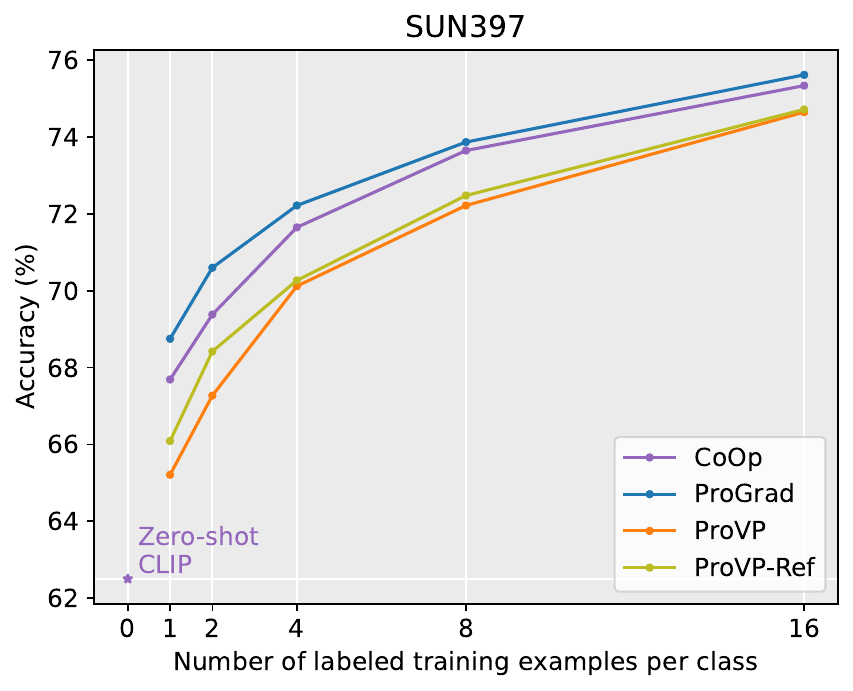}
  \end{subfigure}
  
  \begin{subfigure}[htbp]{0.32\linewidth}
    \includegraphics[width=\linewidth]{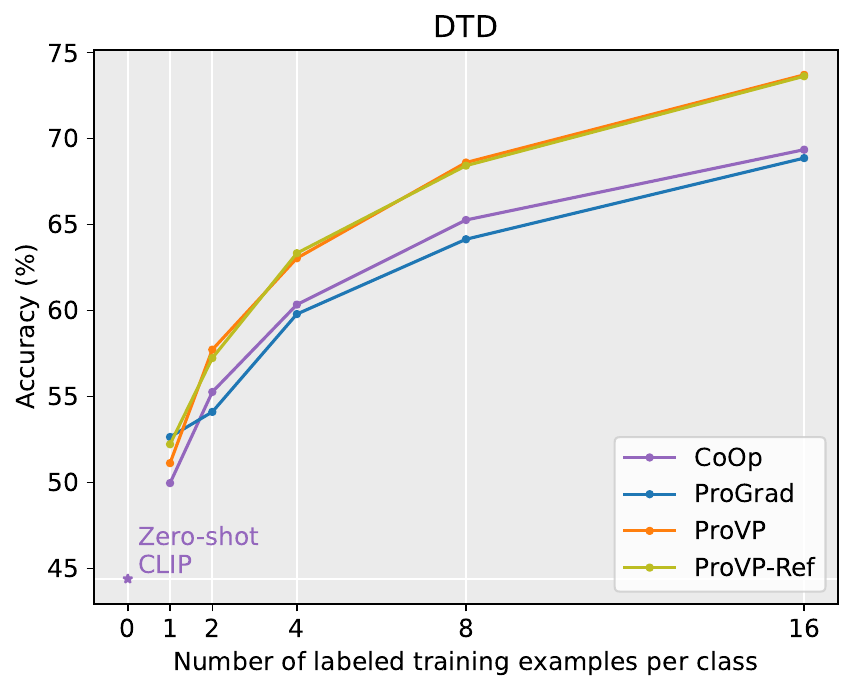}
  \end{subfigure}
  \hfill
  \begin{subfigure}[htbp]{0.32\linewidth}
    \includegraphics[width=\linewidth]{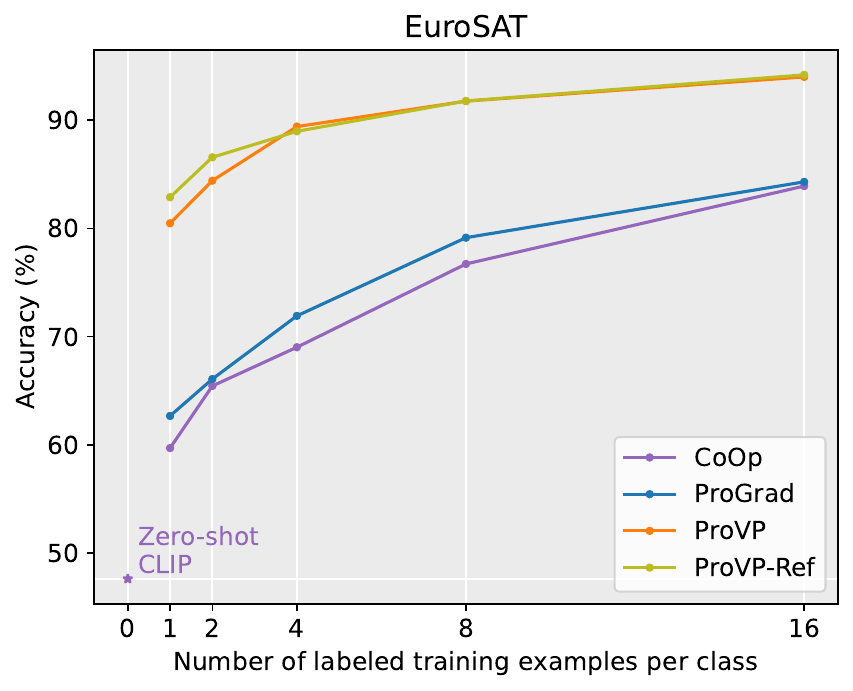}
  \end{subfigure}
  \hfill
  \begin{subfigure}[htbp]{0.32\linewidth}
    \includegraphics[width=\linewidth]{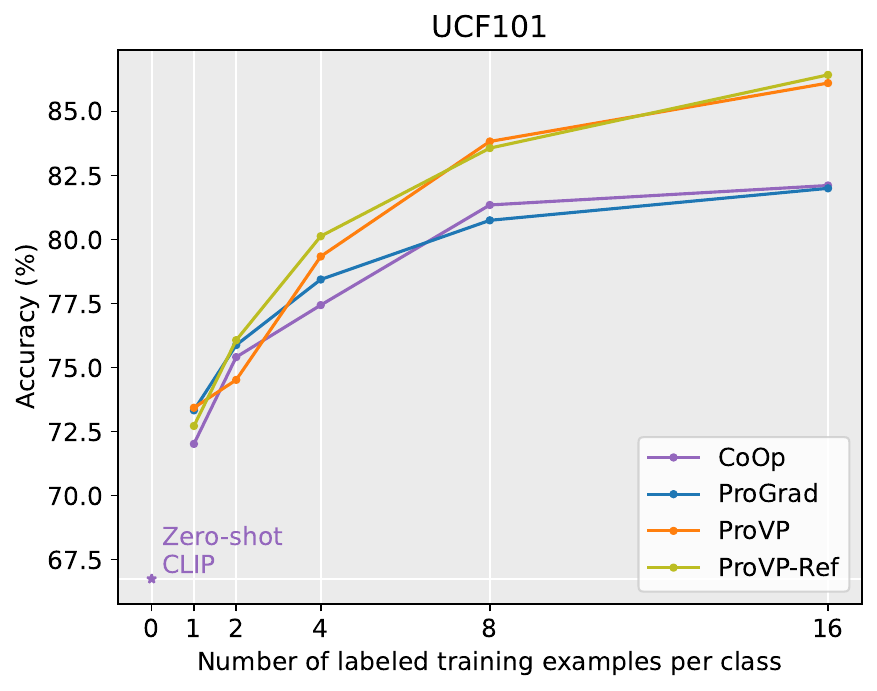}
  \end{subfigure}
  \caption{Full comparison of ProVP-Ref with previous methods on few-shot learning. ProVP-Ref shows significant gains in 7/11 datasets and highly improves the average performance. }
  \label{fig.FewshotResults}
\end{figure*}
\vspace{2mm}
By employing $Loss_{Ref}$ to align the refined features with prompts to the pre-trained distribution, the encoded image embeddings from the tuned image encoder $f_p(\cdot)$ become more recognizable to the text encoder. Consequently, our method can yield prediction results that could be more akin to the zero-shot CLIP, and the release of the strict constraints on the predicted logits allows contrastive feature re-formation to operate more flexibly when conflicts arise between the pre-trained predictions and the downstream ground truth. Benefiting from this strategy, our model as \textbf{ProVP-Ref} can learn a more generalizable representation from pre-trained feature distribution. 
Moreover, our model would be less affected by the conflict of pre-trained knowledge and still adaptable in the downstream task where exists large domain shifts.

\section{Experiments}
\label{sec.Expr}

We evaluate the adaptation and generalization capability of our model in three settings: few-shot learning, base-to-new generalization, and cross-dataset transfer.\\

\noindent\textbf{Few-Shot Learning} is a scenario in which models are trained with a few labeled samples per class to test the adaptation ability of the model when little data is available.\\

\noindent\textbf{Base-to-New Generalization} is used to evaluate the generalizability of the models in a zero-shot manner. The dataset is split into base and new sets with no shared classes. The models are trained on the base sets only but tested on both sets.\\
  
\noindent\textbf{Cross-Dataset Transfer} is used to test the zero-shot transferability of the model. The model is only trained on the source dataset in a few-shot manner and evaluated on various downstream datasets. In this paper, we select ImageNet as the source and test the model on the remaining 10 datasets.\\

\noindent\textbf{Datasets:} In line with previous methods~\citep{CoOp,CoCoOp}, we use a collection of 11 diverse image recognition datasets in all settings. These datasets cover a various spectrum of visual categories, including generic object classification exemplified by ImageNet~\citep{Imagenet} and Caltech101~\citep{Caltech101}, as well as fine-grained classification across FGVCAircraft~\citep{FGVCAircraft}, Flowers102~\citep{OxfordFlowers}, Food101~\citep{food101}, StanfordCars~\citep{StanfordCars} and OxfordPets~\citep{OxfordPets}. Also, we incorporate datasets categorized under actions and scenes such as UCF101~\citep{UCF101} and SUN397 \citep{SUN397}, and datasets for specialized categories like EuroSAT~\citep{EuroSAT} and DTD~\citep{DTD}. In the few-shot learning task, we also follow prior works~\citep{CLIP,CoOp,ProGrad}, and train the model with 1, 2, 4, 8, and 16 shots, testing on full test sets. For base-to-new generalization and cross-dataset transfer, we train the model on the split dataset of 16-shot settings. \\

\noindent\textbf{Training Details:} Our implementation is based on the open-source CLIP~\citep{CLIP} with ViT-B/16~\citep{vit}, and all results are averaged over three runs as prior settings (\citep{CoOp,CoCoOp}). We utilize Xavier\citep{XavierRand} to initialize prompts and fix the progressive decay $\alpha$ as $0.1$. All models are trained with a batch size of 32 and a learning rate of 0.1 with SGD optimizer, while ImageNet and SUN397 benefit from a large learning rate of 5.0. We run all experiments on a single NVIDIA A100 GPU. For few-shot learning, the visual prompt length at each layer is set to 50. The maximum epoch is set to 200 for 16/8 shots, 100 for 4/2 shots, and 50 for 1 shot for all datasets following CoOp~\citep{CoOp}. $\lambda$ is set to 0.1 for ImageNet, StanfordCars, and OxfordFlowers, and 1 for the remaining datasets. For base-to-new generalization, we shorten the prompt length to 16 and the maximum epoch to 100. $\lambda$ is set to 1 except ImageNet to 0.1.

\subsection{Results of Few-Shot Learning}
\label{subsec.fewshot}
\begin{table*}[tbp]
  \footnotesize
  \centering
  
  \caption{Comparison on the base-to-new generalization setting. H: Harmonic mean of base and new performance. Within all prompt-based methods, ProVP-Ref achieves the SOTA results on both base and new classes.}
  \label{tab.Base2newResult}
  \begin{tabular}{l|ccc|ccc|ccc|ccc}
    \toprule
    \multirow{2}{*}{Methods}& \multicolumn{3}{c}{\emph{Average}} & \multicolumn{3}{|c}{ImageNet}
    & \multicolumn{3}{|c}{Caltech101} & \multicolumn{3}{|c}{OxfordPets} \\
    \cmidrule(lr){2-4} \cmidrule(lr){5-7} \cmidrule(lr){8-10} \cmidrule(lr){11-13}
    & Base & New & H & Base & New & H & Base & New & H & Base & New & H \\
      \midrule
      CLIP        & 69.34 & \textbf{74.22} & 71.70 & 72.43 & 68.14 & 70.22
                  & 96.84 & 94.00 & 95.40 & 91.17 & 97.26 & 94.12\\
      CoOp        & 82.69 & 63.22 & 71.66 & \textbf{76.47} & 67.88 & 71.92
                  & 98.00 & 89.81 & 93.73 & 93.67 & 95.29 & 94.47\\
      CoCoOp      & 80.47 & 71.69 & 75.83 & 75.98 & \textbf{70.43} &    \textbf{73.10} 
                  & 97.96 & 93.81 & 95.84 & 95.20 & 97.69 & 96.43\\
      ProGrad     & 81.89 & 71.85 & 76.54 & 76.35 & 69.26 & 72.63
                  & 97.91 & \textbf{94.40} & 96.12 & 94.86 & 97.52 & 96.17\\
      ProDA  & 81.56 & 72.30 & 76.65 & 75.40 & 70.23 & 72.72 
& 98.27 & 93.23 & 95.68 & 95.43 & \textbf{97.83} & 96.62 \\ 
      \midrule
      \rowcolor{gray!15} 
      ProVP       & 85.14 & 69.57 & 76.57 & 75.59 & 69.36 & 72.34 
                  & \textbf{99.01} & 93.34 & 96.09 & 95.80 & 96.18 & 95.99\\
      \rowcolor{gray!15}
      ProVP-Ref  & \textbf{85.20}  & 73.22 & \textbf{78.76} & 75.82 & 69.21 & 72.36     
                  & 98.92 & 94.21 & \textbf{96.51} & \textbf{95.87} & 97.65 & \textbf{96.75}\\
      \bottomrule
      \end{tabular}
  \vspace{1mm}   
  
  \begin{tabular}{l|ccc|ccc|ccc|ccc}
      \toprule
      \multirow{2}{*}{Methods} & \multicolumn{3}{c}{StanfordCars} & 
      \multicolumn{3}{|c}{Flowers102} & \multicolumn{3}{|c}{Food101} & 
      \multicolumn{3}{|c}{FGVCAircraft}\\
      \cmidrule(lr){2-4} \cmidrule(lr){5-7} \cmidrule(lr){8-10} \cmidrule(lr){11-13}
      & Base & New & H & Base & New & H & Base & New & H & Base & New & H \\
      \midrule
      CLIP        & 63.37 & \textbf{74.89} & 68.65 & 72.08 & \textbf{77.80} & 74.83       & 90.10 & 91.22 & 90.66 & 27.19 & 36.29 & 31.09 \\
      CoOp        & 78.12 & 60.40 & 68.13 & 97.60 & 59.67 & 74.06 
                  & 88.33 & 82.26 & 85.19 & 40.44 & 22.30 & 28.75 \\
      CoCoOp      & 70.49 & 73.59 & 72.01 & 94.87 & 71.75 & 81.71 
                  & 90.70 & \textbf{91.29} & 90.99 & 33.41 & 23.71 & 27.74 \\
      ProGrad     & 75.17 & 74.37 & \textbf{74.77} & 95.44 & 74.04 & 
      \textbf{83.39} & \textbf{90.73} 
                  & 91.27 & \textbf{91.00} & 38.88 & 31.63 & 34.88\\
      ProDA       & 74.70 & 71.20 & 72.91 & 97.70 & 68.68 & 80.66 &
                    90.30 & 88.57 & 89.43 & 36.90 & \textbf{34.13} & 35.46 \\

      \midrule
      \rowcolor{gray!15}
      ProVP       & \textbf{80.97} & 63.27 & 71.03 & \textbf{98.45} & 65.39 & 
      78.58       & 90.16 & 90.88 & 90.52 & 46.04 & 25.29 & 32.65\\
      \rowcolor{gray!15}
      ProVP-Ref  & 80.43 & 67.96 & 73.67 & 98.42 & 72.06 & 83.20
                  & 90.32 & 90.91 & 90.61 & \textbf{47.08} & 29.87 & \textbf{36.55}\\
      \bottomrule
    \end{tabular}
  \vspace{1mm}
  
    \begin{tabular}{l|ccc|ccc|ccc|ccc}
      \toprule
      \multirow{2}{*}{Methods} & \multicolumn{3}{c}{SUN397} & \multicolumn{3}{|c}{DTD} & \multicolumn{3}{|c}{EuroSAT} & \multicolumn{3}{|c}{UCF101} \\
      \cmidrule(lr){2-4} \cmidrule(lr){5-7} \cmidrule(lr){8-10} \cmidrule(lr){11-13} 
      &Base&New&H&Base&New&H&Base&New&H&Base&New&H\\
      \midrule
      CLIP        & 69.36 & 75.35 & 72.23 & 53.24 & \textbf{59.90} & 56.37 
                  & 56.48 & 64.05 & 60.03 & 70.53 & \textbf{77.50} & 73.85\\
      CoOp        & 80.60 & 65.89 & 72.51 & 79.44 & 41.18 & 54.24
                  & 92.19 & 54.74 & 68.69 & 84.69 & 56.05 & 67.46\\
      CoCoOp      & 79.74 & 76.86 & 78.27 & 77.01 & 56.00 & 64.85
                  & 87.49 & 60.04 & 71.21 & 82.33 & 73.45 & 77.64\\
      ProGrad     & \textbf{80.85} & 74.93 & 77.78 & 77.16 & 54.63 & 63.97
                  & 88.91 & 53.75 & 67.00 & 84.49 & 74.52 & 79.19\\
      ProDA       & 78.67 & \textbf{76.93} & 77.79 & 80.67 & 56.48 & 66.44 
                  & 83.90 & 66.00 & 73.88 & 85.23 & 71.97 & 78.04 \\

      \midrule
      \rowcolor{gray!15}
      ProVP       & 80.33 & 73.75 & 76.90 & \textbf{84.76} & 52.82 & 65.08
                  & \textbf{97.46} & 63.47 & 76.88 & 87.99 & 71.55 & 78.92\\
      \rowcolor{gray!15}
      ProVP-Ref  & 80.67 & 76.11 & \textbf{78.32} & 83.95 & 59.06 & \textbf{69.34}  
                  & 97.12 & \textbf{72.91} & \textbf{83.29} & \textbf{88.56} & 75.55 & \textbf{81.54}\\
      \bottomrule
    \end{tabular}
\end{table*}

\begin{figure*}[!htbp]
  \centering
  \begin{subfigure}{0.99\textwidth}
  \includegraphics[width=\textwidth]{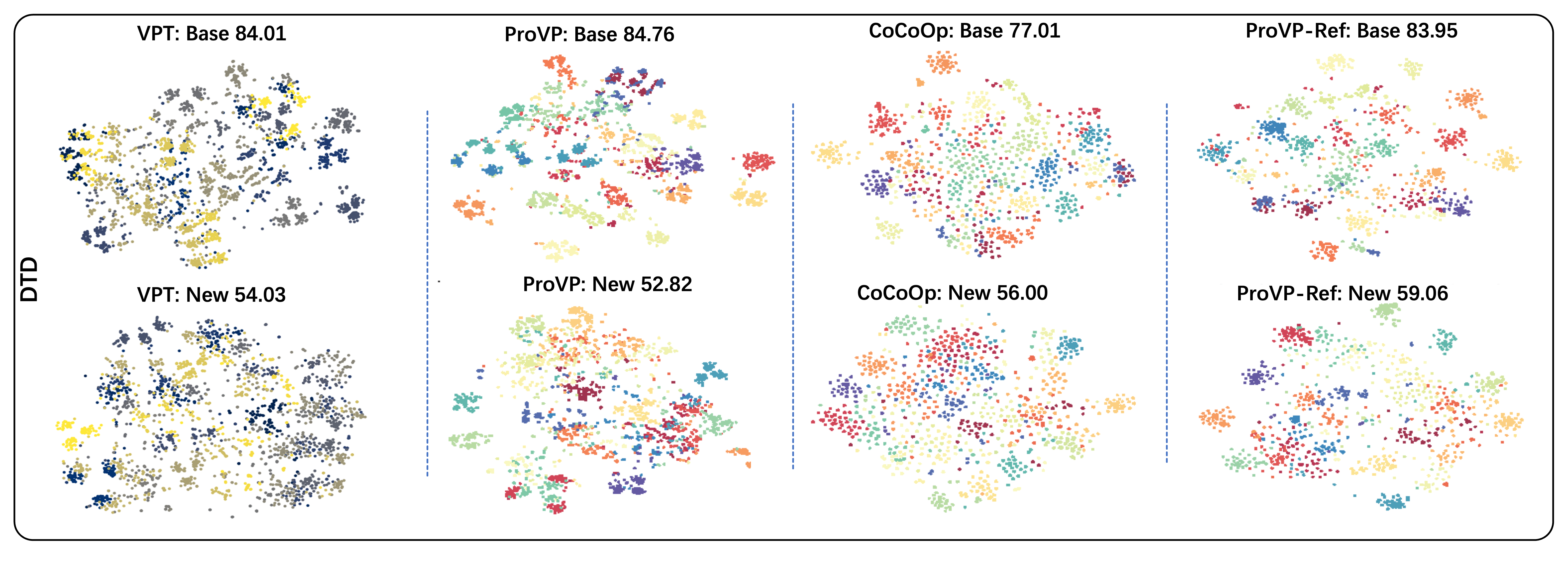}
  \end{subfigure}
  \vspace{2mm}
  \begin{subfigure}{0.99\textwidth}
  \includegraphics[width=\textwidth]{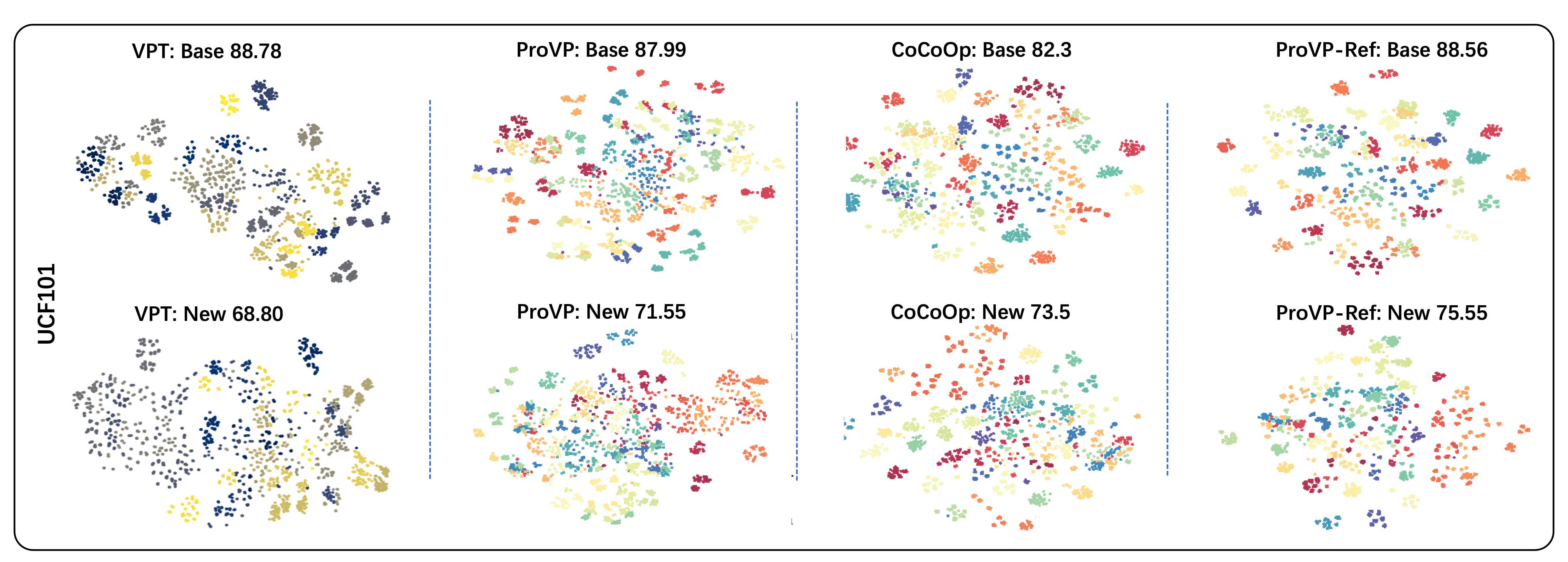}
  \end{subfigure}
  \caption{T-SNE plots of image embeddings in VPT, ProVP, CoCoOp, and ProVP-Ref are presented on two diverse datasets (Noticing that neither CoOp, CoCoOp, nor zero-shot CLIP learns on the image branch. Thus, they share the same image feature visualization). It is evident that the embeddings in ProVP and ProVP-Ref are more separable. Furthermore, since ProVP-Ref has learned from the combination of pre-trained and downstream knowledge, it shares a more similar feature representation to zero-shot CLIP, providing a better generalization capability.}
  \label{fig.tSneResults}
\end{figure*}

The full results of few-shot learning are depicted in in Figure \ref{fig.FewshotResults}. We compare ProVP and ProVP-Ref with two text-based prompt learning methods CoOp~\citep{CoOp} and ProGrad~\citep{ProGrad}.
In terms of the average performance over 11 datasets, ProVP-Ref shows a significant improvement over all previous works in all shot settings. 
Notably, ProVP-Ref gains an absolute improvement as $\mathbf{2.83}$\% compared to the previous SOTA (CoOp) in the 16-shot scenario, evidencing the enhanced adaptation capability of our approach. The performance gap lies in particular datasets that have large domain shifts as EuroSAT and FGVCAircraft, which also confirms that our vision-based prompt learning methods have stronger adaptation ability.  Additionally, comparing these two text prompting methods, we found the average performance of ProGrad lags slightly behind CoOp in the 16-shot setting, we doubt the reason might be that the learning of ProGrad has been negatively impacted by the incorrect prediction of zero-shot CLIP. In contrast, ProVP-Ref shows a consistent improvement with ProVP and contrastive feature re-formation further improves the performance under small shots (e.g., 1-shot). This suggests that ProVP-Ref has better harmonized the maintaining of pre-trained knowledge with the downstream learning process and benefited the model performance when training samples are scarce.

\begin{figure}[t]
  \centering
  \includegraphics[width=0.45\textwidth]{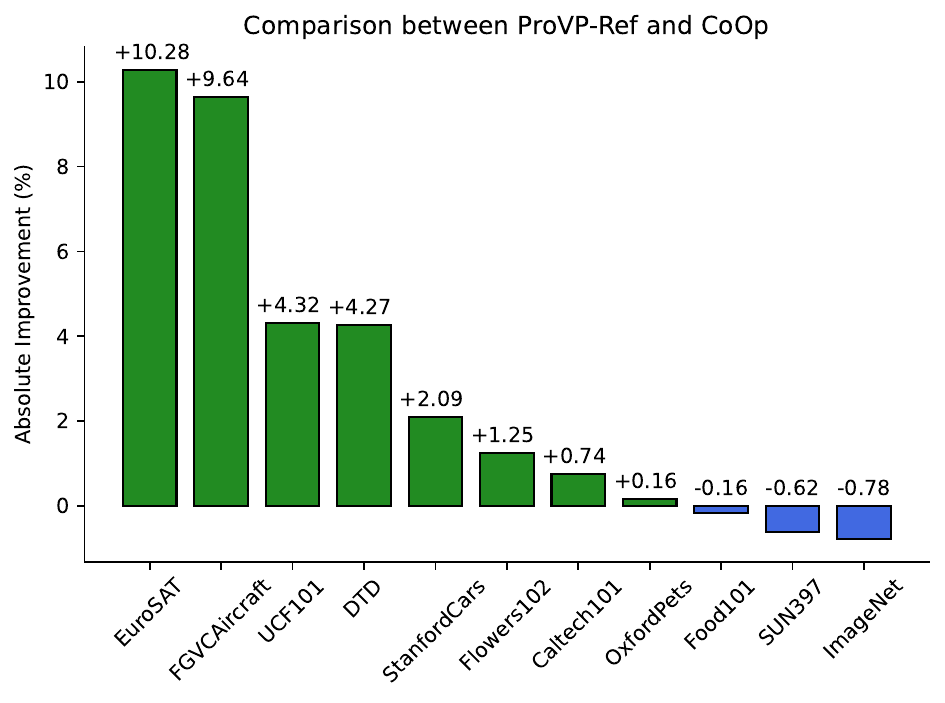}
  \caption{Comparison between ProVP-Ref and CoOp \citep{CoOp} in few-shot learning (16-shot setting).}
  \label{fig.diff-ProVP-CoOp}
\end{figure}

The detailed comparison between our ProVP-Ref and CoOp is displayed in Figure \ref{fig.diff-ProVP-CoOp}. 
ProVP-Ref makes an obvious improvement over those datasets which has large domain shifts from the pre-trained data, 
such as EuroSAT (+10.28\%), FGVCAircraft (+9.64\%), UCF101 (+4.32\%) and DTD (+4.27\%) and also gains improvement on fine-grained datasets as StanfordCars (+2.09\%) and Flowers102 (+1.25\%). On the other hand, we found that ProVP-Ref has obtained less appealing results on two datasets ImageNet and SUN397. 
  A shared characteristic between these two datasets is that they both possess a considerable number of class categories. 
  We have found that this factor probably blocks the performance of our method and explored the reasons which are detailed in the further analysis (see Sec.\ref{subsec.furtherana}).

\begin{table*}[t]
    \footnotesize
    \centering
    \caption{Comparison on cross-dataset transfer benchmark. ProVP-Ref achieves competitive results with the previous SOTA method.}
    
    \begin{tabular*}{\linewidth}{@{\extracolsep{\fill}}l|cccccccccccc}
    \toprule
    & \rotbox{Caltech101} & \rotbox{OxfordPets} & \rotbox{StanfordCars} & \rotbox{Flowers102} & \rotbox{Food101} & \rotbox{FGVCAircraft} & \rotbox{SUN397} & \rotbox{DTD} & \rotbox{EuroSAT} & \rotbox{UCF101} & \rotbox{\emph{Average}} \\
    \midrule
    CLIP &  93.35 & 88.25 & \textbf{65.48} & 67.44 & 83.65 & 23.67 & 62.59 & 44.27 & 42.01 & 65.13 & 63.58 \\
    CoOp &   93.70 & 89.14 & 64.51 & 68.71 & 85.30 & 18.47 & 64.15 & 41.92 & 46.39 & 66.55 & 63.88 \\
    CoCoOp &  \textbf{94.43} & 90.14 & 65.32 & \textbf{71.88} & 86.06 & 22.94 & \textbf{67.36} & 45.73 & 45.37 & \textbf{68.21} & 65.74 \\
    \midrule
    ProVP-Ref  & 93.79	& \textbf{91.58}	& 65.29	& 71.62	& \textbf{86.17}	& \textbf{24.51}	& 66.29& \textbf{45.97} & \textbf{51.95} & 67.72 & \textbf{66.49}\\
    \bottomrule
    \end{tabular*}
    \label{tab.crossdataset}
\end{table*}

\subsection{Results of Base-to-New Generalization}
\label{subsec.base2new}
When testing on base-to-new generalization, we evaluate the model generalizability in a zero-shot setting. The dataset is split into base and new sets with no shared classes and the models are trained on the base set only but tested on both base and new sets.

The overall results are shown in Table \ref{tab.Base2newResult}. We have three key findings: 1) ProVP shows strong adaptation capability. In terms of the base performance, ProVP achieves SOTA results on average over 11 datasets with an absolute gain of 2.45\% over the previous best method (CoOp). Great improvements can be seen at FGCVAircraft (+6.64\%), EuroSAT (+5.27\%), DTD (+5.32\%), and UCF101 (+3.30\%). Despite the performance drops in new classes, ProVP still outperforms all previous works on the average Harmonic mean evaluation, which claims that the gains on the base set still outweigh the decline. 2) Contrastive Feature Re-formation highly alleviates the generalization deterioration problem. Although ProVP shows impressive results on the base set, the weaker performance on new classes highlights the persistence of generalization challenges. By combining contrastive feature re-formation, ProVP-Ref improves the new performance significantly with an absolute \textbf{3.53}\% improvement over ProVP and also surpasses the previous SOTA text-based prompt learning method (ProGrad) by \textbf{1.37}\%. A clear performance boost can be observed on all datasets compared to ProVP. For a more tangible illustration of our method's effectiveness, we refer to Figure \ref{fig.tSneResults} for the visualization of image embeddings of different methods. It is evident that ProVP-Ref maintains a feature representation more closely ali-gned with the zero-shot CLIP model while also ensuring greater separability than text-based approaches. This alignment is in harmony with the observed improvements in both model adaptation and generalization capabilities of our approach. 3) ProVP-Ref represents a win-win scenario for both base and new class recognition. Observing all the results in Table \ref{tab.Base2newResult}, we found the previous methods usually sacrifice the adaptation capability (base performance) of the models, maintaining more pre-trained knowledge to improve generalization capability (new performance). On the contrary, ProVP-Ref obtains state-of-the-art results on both base
and new performance simultaneously, giving the highest Harmonic mean. This indicates that our approach preserves more task-relevant general knowledge without conflicting with the learning process, thereby enhancing performance for both base and new classes and providing the best trade-off for adaptation and generalization.
For further discussion, ProVP-Ref gave a sub-optimal new performance on particular datasets such as StanfordCars, Flowers102, and FGVCAircraft.  Nevertheless, we found all these datasets have a common characteristic: the best performance on new classes is still achieved by zero-shot CLIP while its base is far behind all other methods. One possible reason is the pre-trained knowledge of CLIP, which contributes to the new performance, may heavily conflict with the downstream tasks on such datasets and may be degraded after the tuning. It is worth noting that this problem exists in all prompt learning methods on these datasets, and despite this issue, ProVP-Ref still achieves a competitive Harmonic mean compared with other prompt-based methods.

\begin{table*}[tbp]
    \centering
    \caption{Results for $\alpha$ and $\lambda$ ablation study. Table \ref{tab.AblationOnAlpha} ablates $\alpha$ in few-shot learning on 16-shot setting, and Table \ref{tab.AblationOnLam} and \ref{tab.AblationOnLam2} are ablation studies on few-shot learning and base-to-new generalization, respectively.}
\begin{subtable}[t]{0.48\textwidth}
      \centering
      \caption{Different $\alpha$ in few-shot learning. }
      \begin{tabular*}{\linewidth}{@{\extracolsep{\fill}}lccc}
        \toprule
        $\alpha$       & Caltech101   &  DTD & StanfordCars   \\
        \midrule
        0.01           & 96.55 & 73.29 &  82.88 \\ 
        0.1            & \textbf{96.63} &  \textbf{73.62} & \textbf{84.85} \\
        0.3            & 96.63 & 72.48 & 83.43 \\
        0.5            & 96.39 & 70.57  & 80.05 \\
        0.7            & 95.53 & 67.06 & 75.63 \\
        0.9            & 95.27 & 61.96  & 69.04 \\
        \bottomrule
        \end{tabular*}
      \label{tab.AblationOnAlpha}
\end{subtable}
\hfill
\begin{subtable}[t]{0.48\textwidth}
      \centering
      \caption{Different $\lambda$ in few-shot learning. }
     \begin{tabular*}{\textwidth}{@{\extracolsep{\fill}}lccc}
     \toprule
     $\lambda$        & Caltech101 & DTD  & StanfordCars  \\
     \midrule
     0.     & 96.50 & 73.70 & 84.68 \\
     0.2   & 96.63 & \textbf{73.80} & 84.70  \\
     0.4     & \textbf{96.73} & 73.70 & \textbf{84.71} \\
     0.6     & 96.73 & 73.60 & 84.53 \\
     0.8     & 96.62 & 73.68 & 84.55 \\
     1.0     & 96.63 & 73.62 & 84.64 \\
    \bottomrule
    \end{tabular*}
    \label{tab.AblationOnLam}
  \end{subtable}

    \vspace{1mm}
    \begin{subtable}[t]{\linewidth}
    \centering 
    \caption{Different $\lambda$ in base-to-new generalization.}
    \begin{tabular*}{\linewidth}{@{\extracolsep{\fill}}ccccccc}
      \toprule
      $\lambda$& \multicolumn{3}{c}{DTD}& \multicolumn{3}{c}{StanfordCars}\\
      \midrule
            & Base  & New & H     & Base  & New & H     \\
      \cmidrule{2-4}
      \cmidrule{5-7}
      0     & \textbf{84.76} & 52.82 & 65.08 & 80.97 & 63.27 & 71.03 \\
      0.2   & 84.61 & 55.03 & 66.69 & 81.07 & 64.29 & 71.71 \\
      0.4   & 84.53 & 55.80 & 67.22 & \textbf{81.21} & 66.15 & 72.91 \\
      0.6   & 84.18 & 58.01 & 68.69 & 80.78 & 66.97 & 73.23 \\
      0.8   & 83.95 & 58.82 & 69.17 & 80.36 & \textbf{68.07} & \textbf{73.71} \\
      1.0   & 83.95 & \textbf{59.06} & \textbf{69.34} & 80.43 & 67.96 & 73.67 \\
      \bottomrule
      \end{tabular*}
    \label{tab.AblationOnLam2}
\end{subtable}
\end{table*}

\begin{figure*}[tbp]
    \centering
    \includegraphics[width=0.8\linewidth]{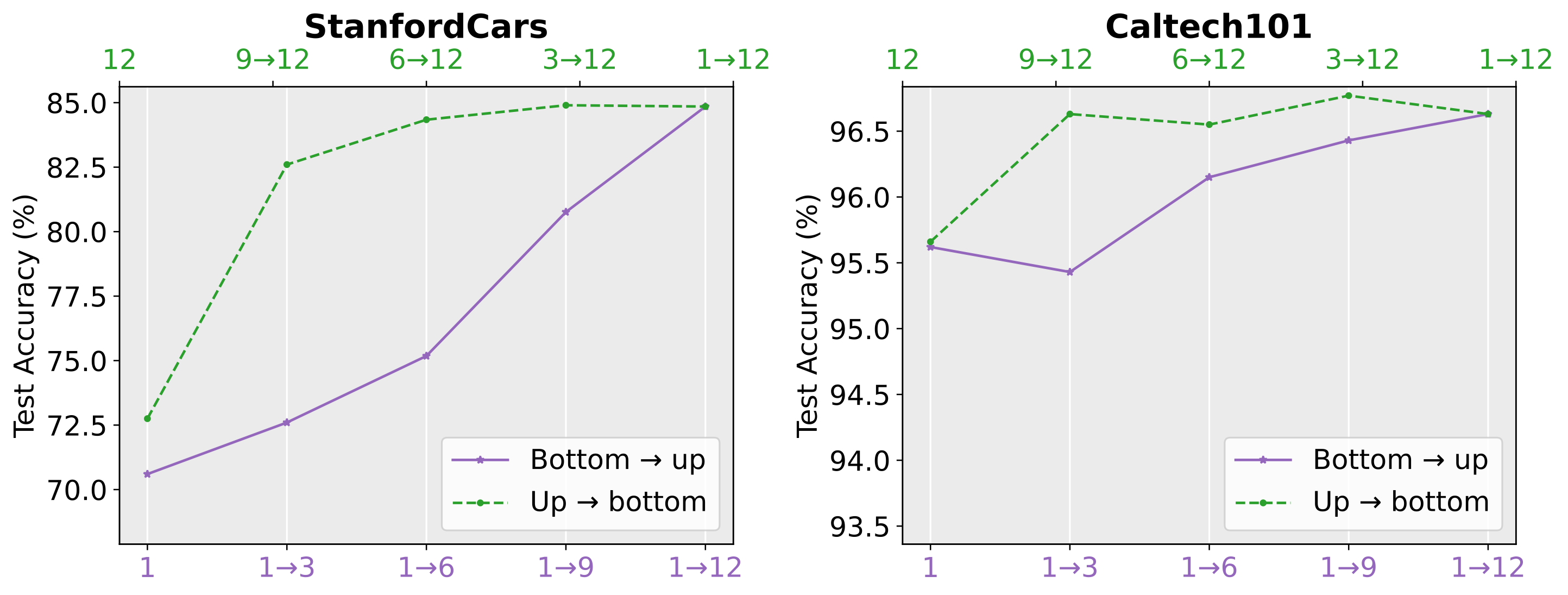}
    \caption{Ablation on prompt depth. $i\to j$ means the range of the transformer layers inserted with prompts. { 'bottom$\to$up': learnable prompts are systematically inserted starting from the first network layer close to the input side (Layer 1), 'up$\to$bottom': prompts are consistently inserted up to the final layer (Layer 12)}}
    \label{fig.Depth}
  \end{figure*}
\begin{table*}[t]
    \centering
    \caption{ProVP-Ref vs. text prompting with larger prompt length. $x$ in model$_x$: the total prompt length in the model.  Few-shot$-Acc.$: the average result of the 16-shot learning on the 11 datasets.} 
    \begin{tabular*}{\linewidth}{@{\extracolsep{\fill}}lccccc}
         \toprule
         Models & \#Parameters & Few-shot$-Acc.$ & Base & 
         New & H \\
         \midrule
         CoOp$_{60}$       & 30720 & 80.88 & 83.23 & 68.37 & 75.07 \\
         CoCoOp$_{4}$      & 34846 & /     & 80.43 & 71.69 & 75.83 \\
         ProGrad$_{60}$    & 30720 & 80.47 & 83.05 & 71.00 & 76.55 \\
         \midrule
        
         ProVP-Ref$_{48}$  & 36864 & 81.72 & 84.48 & 72.14 & 77.83\\
         ProVP-Ref$_{60}$  & 46080 & \textbf{81.79} & \textbf{84.61} & \textbf{72.91} & \textbf{78.32}\\
         \bottomrule
    \end{tabular*}
    \label{tab.ParaComp}  
\end{table*}

\subsection{Cross-Dataset Transfer} 
The overall results for cross-dataset transfer are presented in Table~\ref{tab.crossdataset}. ProVP-Ref achieves competitive results on this benchmark compared to previous SOTA and achieves the best performance on 5 datasets and the overall average  on all datasets. Notably, our method shows better performance towards datasets with large domain gaps (e.g., 5.56\% for EuroSAT), indicating that ProVP-Ref indeed learns a more generalizable representation for the visual prompt learning and is more robust to the domain shifts.

\subsection{Ablation Study}
\label{sec.Ablation}

\noindent\textbf{The choice of $\mathbf{\alpha}$.} The ablation results of $\alpha$ are shown in Table \ref{tab.AblationOnAlpha}. We tested different $\alpha$ on three types of downstream datasets as Caltech101 \citep{Caltech101} for general objects recognition, StanfordCars \citep{StanfordCars} for fine-grained classification, and DTD \citep{DTD}, which have a large domain shift.  As $\alpha$ increases, the behavior of ProVP becomes closer to VPT-Shallow: the performance is still promising on Caltech101 but degrades on DTD and StanfordCars. It suggests that the adaptation ability is still determined by the newly inserted prompts.\\

\noindent\textbf{The Effect of  $Loss_{Ref}$ weight $\lambda$.} 
To further investigate the impact of contrastive feature re-formation, we conduct an ablation study to analyze the effect of $Loss_{Ref}$ in Eq.~\ref{eq.ReformatLoss}. The results of the study for $\lambda$ on base-to-new generalization are presented in Table~\ref{tab.AblationOnLam2}. As $\lambda$ increases, the new performance is consistently enhanced across both datasets, confirming the efficacy of our approach. Notably, an appropriately chosen $\lambda$ can also improve the base accuracy, indicating that preserving knowledge through feature space offers adaptability for visual prompts. For DTD, we suspect that the degradation in base performance is due to the large gap between it and CLIP.\\

\noindent\textbf{Prompt Depth.} Figure~\ref{fig.Depth} shows the ablation study on the depth of layers with prompts. { In this ablation, we investigate two distinct settings: the 'bottom$\to$up' and 'up$\to$bottom' configurations. In the 'bottom$\to$up' setting, learnable prompts are systematically inserted starting from the first network layer (Layer 1), with the number of layers for prompt insertion being adjusted. Conversely, in the 'up$\to$bottom' setting, prompts are consistently inserted up to the final layer (Layer 12). } Consistent with prior findings~\citep{VPT}, there is a positive correlation between depth and overall performance. Regarding the optimal placement of prompts, ProVP-Ref derives greater benefits from prompts near the output layers, which holds for both general recognition such as Caltech101, and fine-grained datasets like StanfordCars. Interestingly, this diverges from the pattern observed in VPT, which emphasizes the importance of prompts at shallow layers. { We speculate that this difference may stem from the choice of backbone models. While VPT experiments are conducted using ViT~\citep{vit}, our approach is built upon CLIP~\citep{CLIP}. Given that CLIP is pre-trained on a large corpus of image-text pairs, the bottom layers may have already captured more sufficient general representations of input images than ViT models. In contrast, the 'up' layers, closer to the output side, may prioritize capturing task-specific information relevant to downstream tasks. Despite the pre-training, these layers in CLIP are not specifically trained for classification tasks, leaving room for optimization. Consequently, fine-tuning these 'up' layers in CLIP could bolster the model's comprehension of downstream tasks, potentially resulting in enhanced performance.}\\
\begin{table}[t]
      \centering
      \caption{ProVP vs.  VPT in base-to-new.}
      \begin{tabular*}{\linewidth}{@{\extracolsep{\fill}}lccc}
        \toprule
        Methods         & Base  & New &   H   \\
        \midrule
        VPT-Shallow     & 80.06 & 72.35 & 76.01 \\
        VPT-Sha-Ref & 79.45 & 72.70 & 75.93 \\
        VPT-Deep        & 84.97 & 68.74 & 76.00 \\
        VPT-Deep-Ref    & 84.85 & 71.50 & 77.60 \\
        \midrule
         ProVP           & 85.14 & 69.57 & 76.57 \\
         ProVP-Ref       & \textbf{85.20}  & \textbf{73.22} & \textbf{78.76}\\
        \bottomrule
        \end{tabular*}
      \label{tab.VPTvsProVP}
\end{table}
\begin{figure}[t]
    \centering
    \includegraphics[scale=0.45]{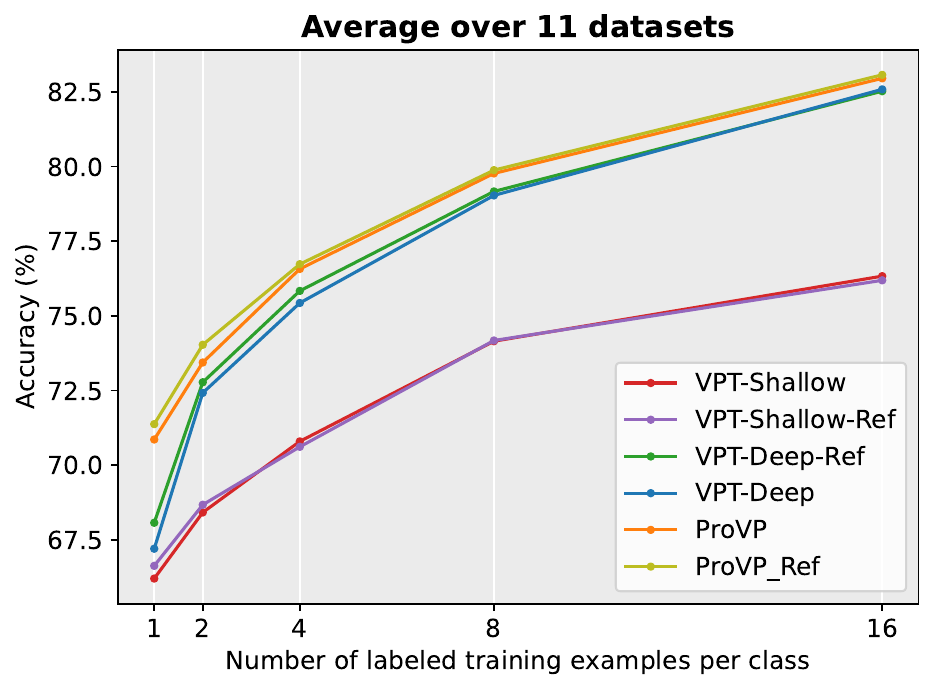}
    \caption{ProVP vs.  VPT on few-shot learning. With the same hyperparameters, ProVP shows a consistently better performance compared to VPT variants.}
    \label{fig.ProVPvsDeep} 
\end{figure}
\begin{table}[t]
      \centering
      \caption{Further analysis of contrastive feature re-formation. ProVP-KD combines ProVP with conventional knowledge distillation, and ProVP-PA utilizes the prompt-aligned strategy proposed in \citep{ProGrad}. Base, New, and H : the average results on 11 datasets.}
      \begin{tabular*}{\linewidth}{@{\extracolsep{\fill}}lccc}
        \toprule
        Methods      & Base  & New &   H   \\
        \midrule
        ProVP        & 85.14 & 69.57 & 76.57\\
        ProVP-KD     & 82.19 & \textbf{73.40} & 77.55\\
        ProVP-PA     & 84.83 & 72.04 & 77.92 \\
        ProVP-Ref    & \textbf{85.20} & 73.22 & \textbf{78.76}\\
        \bottomrule
        \end{tabular*}
      \label{tab.furtherbase2new}
\end{table} 

\noindent\textbf{Ablation of Parameters.}
As our visual prompting method utilizes more learnable prompts compared to text-based methods, it is reasonable to question whether the improvements are solely due to the increased parameter size. To address this concern, we compare all open-source methods under similar parameter quantities. The comparison results are presented in Table~\ref{tab.ParaComp}. Despite the reduction in the number of trainable parameters, our method still achieves superior performance on both base and new categories, providing convincing evidence that the enhancements of our methods cannot be solely attributed to larger parameters.

\begin{figure}[t]
    \centering
    \includegraphics[width=0.48\textwidth]{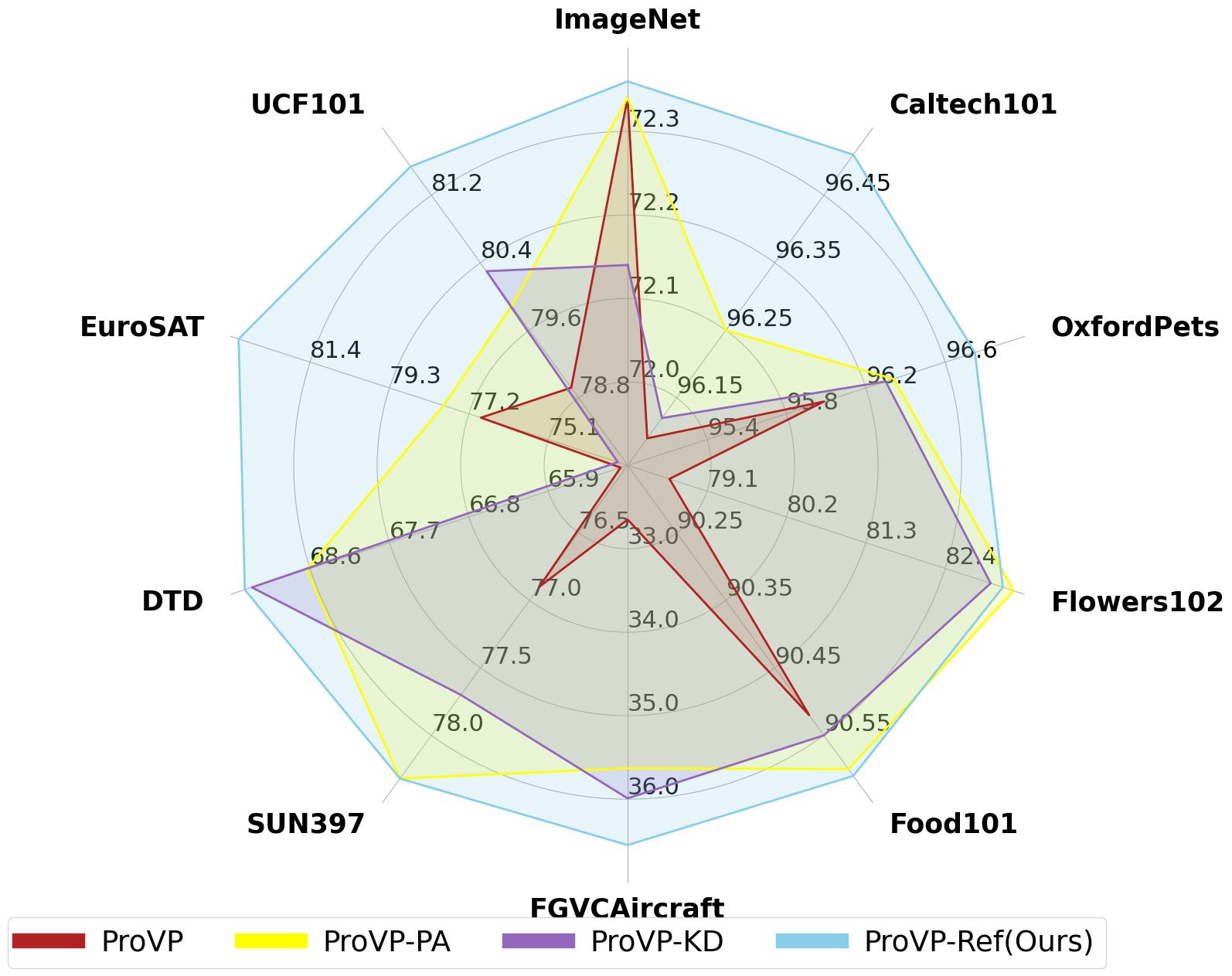}
    \caption{Contrastive feature re-formation vs. other methods when both applied on ProVP. For each dataset, we show in this figure the Harmonic mean performance of the four methods, respectively. ProVP-Ref has achieved the best results compared to the other methods with similar attempts, which also indicates that our method is more effective when adapting visual prompts to downstream tasks. }
    \label{fig.contrastivefeatComp}
\end{figure}
{\subsection{Further Analysis}}
\label{subsec.furtherana}

\noindent\textbf{ProVP vs.  VPT.} 
A thorough comparison of ProVP and VPT structures is presented in Figure \ref{fig.ProVPvsDeep} for few-shot learning and Table \ref{tab.VPTvsProVP} for base-to-new generalization. For few-shot learning, ProVP consistently outperforms VPT across all shot settings and shows a remarkable improvement on extreme low shot settings (e.g., in 1-shot, ProVP outperforms VPT-Deep by \textbf{4.17}\%). Furthermore, in base-to-new generalization, the significant improvement in new performance (+1.72\% on average) also indicates the effectiveness of the instance-specific approach used in ProVP, which further improves the generalization capability.

Moreover, as we emphasized that ProVP could alleviate the oscillation of performance and sensitivity to hyper-parameters that existed in VPT-Deep,  this observation is substantiated by Figure \ref{fig.train}, which depicts the test accuracy of VPT-Deep and ProVP over the prompt learning process. Clearly, VPT-Deep encounters frequent and severe performance drops, occasionally approaching zero. In contrast, with the same hyperparameters, the accuracy curve of ProVP maintains a considerably smoother trajectory. These results affirm that ProVP achieves a more stable tuning compared to the original deep structure.\\

\noindent\textbf{Contrastive Feature Re-formation is more Effective for Visual Prompts.} In Section \ref{sec:method} we have discussed the difference of our contrastive feature re-formation from previous methods as we preserve the pre-trained knowledge through the \textit{feature} space. Although \citep{ProGrad} have shown the methods applied in incremental learning are less efficient for prompt learning, we still evaluated our ProVP with the prompt-aligned method proposed in \citep{ProGrad} and the conventional knowledge distillation \citep{LearnWoForget}. Specifically, \textbf{ProVP-PA} utilizes the method in \citep{ProGrad} to align the gradient to not conflict with the zero-shot predictions of CLIP, and \textbf{ProVP-KD} modifies the training loss as :
\begin{equation}
\centering
 Loss_{KD}=Loss_{ce}+\beta*Loss_{KL},
\end{equation}
where $Loss_{KL}$ was first proposed in \citep{LearnWoForget} and calculates the KL-divergence of the predicted logits of the tuned models as $p(t_i|x)$ and pre-trained zero-shot models $p_{zs}(w_i|x)$ as:
\begin{equation}
    \centering
    Loss_{KL}(x)=-\sum\limits_i p_{zs}(w_i|x)\log \frac{p(t_i|x)}{p_{zs}(w_i|x)}.
\end{equation}

 In our experiment, $\beta$ is set to 1 which is equal to our $\lambda$ setting. The results of such methods are presented in Table~\ref{tab.furtherbase2new}. For ProVP-PA, ProVP-Ref directly shows consistently better performance on both base and new performance. Although ProVP-KD has obtained a marginally better result on the performance of new categories (only +0.18\%), the heavily degraded performance on base classes (\textbf{-3.01\%}) shows that the slightly better generalization capability is likely to be a compromise of the insufficiency of downstream learning, and the gain on new classes cannot compensate for the loss of adaptation capability. \\

\begin{figure}[tbp]
    \centering
    \includegraphics[width=0.45\textwidth]{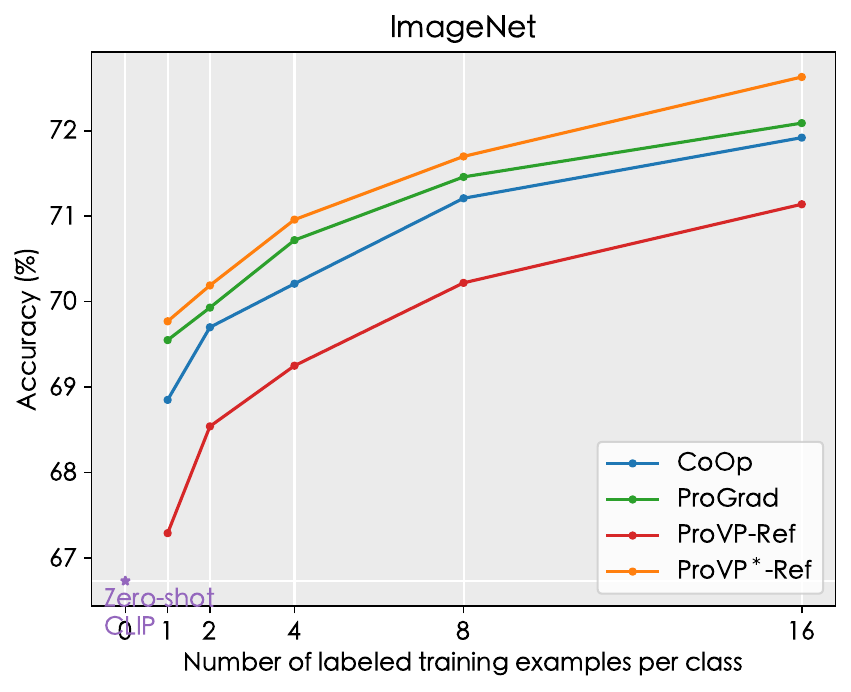}
    \caption{Comparison of ProVP$^*$-Ref, ProVP-Ref, and text-prompting method in the few-shot learning on the ImageNet dataset.}
    \label{fig.FSImgNet}
\end{figure}

{\noindent\textbf{Multi-modal Version of ProVP-Ref.}}
Despite that ProVP-Ref has shown great improvement for most downstream tasks on both adaptation (few-shot learning) and generalization (base-to-new generalization) capability. We found in particular two datasets ImageNet and SUN397, ProVP-Ref obtains less appealing results compared to the text-based prompt learning method. We found the sub-optimal results might be caused by the intrinsic drawback of CLIP \citep{CLIP}: in zero-shot recognition, the text features of CLIP have low distinguishability between different classes, and the distinguishability gets worse when the number of classes increases \citep{OpennessOfCLIP} (e.g., 1000 for ImageNet and 397 for SUN397). As we froze the text branch and leveraged such text features directly, the ambiguous text features may confuse the training of visual prompts, leading to a sub-optimal result.

To overcome such problems, we tried to extend our method by replacing the hand-crafted prompts with the learned text embeddings, which was verified in \citep{OpennessOfCLIP} to be able to alleviate the low distinguishability problem. This extension, denoted as ProVP$^*$-Ref, significantly improves the performance on the ImageNet dataset (+1.49\%) and SUN397 (+1.46\%). { The few-shot results for ImageNet and overall performance are illustrated in Figure \ref{fig.FSImgNet} and Table \ref{tab.VLPrompts}, respectively. These results demonstrate consistent improvements achieved by our ProVP$^*$-Ref in the few-shot learning task compared to previous methods. Additionally, we evaluate ProVP$^*$-Ref in base-to-new generalization, with the results presented in Table \ref{tab.multi-modal-prompt}. Given recent advancements in text-based prompting methods, particularly in maintaining pre-trained knowledge using approaches like LASP \citep{LASP}, we explore integrating such methods with our ProVP$^*$-Ref. This extended approach, named ProVP$^*$-MulRef, leverages progressive prompt learning in both the text and image encoders of CLIP, incorporating LASP for the text branch and contrastive feature re-formation for the image branch to preserve general knowledge. Encouragingly, ProVP$^*$-Ref achieves the best balance of adaptation and generalization capability, with an average Harmonic mean reaching \textbf{79.19}\%. Our findings suggest the potential of multi-modal prompt learning as a promising direction for future research.

\begin{table}[t]
  \footnotesize
  \centering
  \caption{Results of ProVP$^*$-Ref in few-shot learning (16 shots). ProVP$^*$-Ref significantly improves the performance on ImageNet and SUN397 which are limited in ProVP-Ref and also achieves the best average.}
  \begin{tabular*}{\linewidth}{@{\extracolsep{\fill}}l|cccc}
    \toprule
    Method&\multicolumn{3}{c}{\emph{Accuracy.}}\\
    \cmidrule(lr){2-4}
    &ImageNet&SUN397& \emph{Average}\\
    \midrule
    CoOp        & 71.92 & 75.34  & 80.24\\
    ProGrad     & 72.09 & 75.62  & 79.84\\
    \midrule
    ProVP-Ref      & 71.14 & 74.72 & 83.07\\
    ProVP$^*$-Ref  & \textbf{72.63}  & \textbf{76.18} & \textbf{83.41}\\
    \bottomrule
    \end{tabular*}
  \label{tab.VLPrompts}
\end{table}

\section{ Conclusion and Future Work}

In this paper, we introduced a novel progressive structure, ProVP, to address the challenge of visual prompt learning in vision-language models (VLMs). Our approach leverages an instance-specific strategy to enhance the robustness of representations and stabilize the training process. Additionally, we implemented a contrastive feature re-formation technique to mitigate the generalization deterioration that significantly impacts the adaptability of tuned models.

The empirical results demonstrate that ProVP-Ref surpasses all previous single-modal prompt learning methods, showcasing enhanced adaptation and generalization capabilities across three widely used benchmarks. Our method's superiority is particularly notable in datasets experiencing larger domain shifts. However, we observed that ProVP-Ref might face limitations due to the inadequate distinguishability of CLIP text features in certain contexts. To address this, we proposed an extended multi-modal version, ProVP$^*$-Ref, and validated its effectiveness.

We hope that the insights and findings from our study will motivate further research into VLM prompt learning across various image-related tasks \citep{CoOp, xu2023dpl} and video-related tasks \citep{dualdetr, ema, zeroi2v}.

\begin{table}[tbp]
  \centering
  \caption{Comparison of prompt learning methods on single and multi modalities in base-to-new generalization.}
  \label{tab.multi-modal-prompt}
    \begin{tabular}{lcccc}
    \toprule
    Methods   &  Base  & New &   H   \\
    \midrule
    \multicolumn{2}{l}{\textit{Single-modal Methods}}\\
    ProGrad &  81.89 & 71.85 & 76.54 \\
    LASP &82.70 &74.90 &78.61 \\
    ProVP-Ref & 85.20 & 73.22 & 78.76 \\
    \midrule
    \multicolumn{2}{l}{\textit{Multi-modal Methods}}\\
    MaPLe   & 82.28 & \textbf{75.14} & 78.55\\
    ProVP$^*$-Ref   & 84.31 & 74.33 & 79.00 \\
    ProVP$^*$-MulRef  & \textbf{84.49} & 74.81  & \textbf{79.36} \\
    
    \bottomrule
    \end{tabular}
\end{table}

\paragraph{\bf Acknowledgement.} 
 This work is supported by the National Key R$\&$D Program of China (No. 2022ZD0160 900), the National Natural Science Foundation of China (No. 62076119, No. 61921006), and the Collaborative Innovation Center of Novel Software Technology and Industrialization.

\paragraph{\bf Data Availability.} 
This manuscript develops its me-thod based on the publicly available datasets: ImageNet, Caltech101, FGVCAircraft, Flowers102, Food101, StanfordCars, OxfordPets, UCF101, SUN397, EuroSAT and DTD. There is no specific associated data with this manuscript.


\bibliographystyle{spbasic}      

\bibliography{ijcvbib}

\begin{thebibliography}{63}
\providecommand{\natexlab}[1]{#1}
\providecommand{\url}[1]{{#1}}
\providecommand{\urlprefix}{URL }
\expandafter\ifx\csname urlstyle\endcsname\relax
  \providecommand{\doi}[1]{DOI~\discretionary{}{}{}#1}\else
  \providecommand{\doi}{DOI~\discretionary{}{}{}\begingroup \urlstyle{rm}\Url}\fi
\providecommand{\eprint}[2][]{\url{#2}}

\bibitem[{Bahng et~al(2022)Bahng, Jahanian, Sankaranarayanan, and Isola}]{bahng2022exploring}
Bahng H, Jahanian A, Sankaranarayanan S, Isola P (2022) Exploring visual prompts for adapting large-scale models. arXiv preprint arXiv:220317274

\bibitem[{Bossard et~al(2014)Bossard, Guillaumin, and Gool}]{food101}
Bossard L, Guillaumin M, Gool LV (2014) Food-101--mining discriminative components with random forests. In: ECCV, pp 446--461

\bibitem[{Bulat and Tzimiropoulos(2023)}]{LASP}
Bulat A, Tzimiropoulos G (2023) Lasp: Text-to-text optimization for language-aware soft prompting of vision \& language models. \eprint{2210.01115}

\bibitem[{Cimpoi et~al(2014)Cimpoi, Maji, Kokkinos, Mohamed, and Vedaldi}]{DTD}
Cimpoi M, Maji S, Kokkinos I, Mohamed S, Vedaldi A (2014) Describing textures in the wild. In: CVPR, pp 3606--3613

\bibitem[{Conde and Turgutlu(2021)}]{conde2021clip}
Conde MV, Turgutlu K (2021) Clip-art: contrastive pre-training for fine-grained art classification. In: CVPR, pp 3956--3960

\bibitem[{Deng et~al(2009)Deng, Dong, Socher, Li, Li, and Fei-Fei}]{Imagenet}
Deng J, Dong W, Socher R, Li LJ, Li K, Fei-Fei L (2009) Imagenet: A large-scale hierarchical image database. In: CVPR, pp 248--255

\bibitem[{Dosovitskiy et~al(2021)Dosovitskiy, Beyer, Kolesnikov, Weissenborn, Zhai, Unterthiner, Dehghani, Minderer, Heigold, Gelly et~al}]{vit}
Dosovitskiy A, Beyer L, Kolesnikov A, Weissenborn D, Zhai X, Unterthiner T, Dehghani M, Minderer M, Heigold G, Gelly S, et~al (2021) An image is worth 16x16 words: Transformers for image recognition at scale. In: ICLR

\bibitem[{Du et~al(2022)Du, Wei, Zhang, Shi, Gao, and Li}]{du2022learning}
Du Y, Wei F, Zhang Z, Shi M, Gao Y, Li G (2022) Learning to prompt for open-vocabulary object detection with vision-language model. In: CVPR, pp 14084--14093

\bibitem[{Fei-Fei et~al(2004)Fei-Fei, Fergus, and Perona}]{Caltech101}
Fei-Fei L, Fergus R, Perona P (2004) Learning generative visual models from few training examples: An incremental bayesian approach tested on 101 object categories. In: CVPR-W, pp 178--178

\bibitem[{Gao et~al(2021)Gao, Geng, Zhang, Ma, Fang, Zhang, Li, and Qiao}]{gao2021clip}
Gao P, Geng S, Zhang R, Ma T, Fang R, Zhang Y, Li H, Qiao Y (2021) Clip-adapter: Better vision-language models with feature adapters. arXiv preprint arXiv:211004544

\bibitem[{Ghiasi et~al(2022)Ghiasi, Gu, Cui, and Lin}]{ghiasi2022scaling}
Ghiasi G, Gu X, Cui Y, Lin TY (2022) Scaling open-vocabulary image segmentation with image-level labels. In: ECCV, pp 540--557

\bibitem[{Glorot and Bengio(2010)}]{XavierRand}
Glorot X, Bengio Y (2010) Understanding the difficulty of training deep feedforward neural networks. In: AISTATS, pp 249--256

\bibitem[{Gu et~al(2021)Gu, Lin, Kuo, and Cui}]{gu2021open}
Gu X, Lin TY, Kuo W, Cui Y (2021) Open-vocabulary object detection via vision and language knowledge distillation. In: ICLR

\bibitem[{He et~al(2016)He, Zhang, Ren, and Sun}]{resnet}
He K, Zhang X, Ren S, Sun J (2016) Deep residual learning for image recognition. In: CVPR, pp 770--778

\bibitem[{Helber et~al(2019)Helber, Bischke, Dengel, and Borth}]{EuroSAT}
Helber P, Bischke B, Dengel A, Borth D (2019) Eurosat: A novel dataset and deep learning benchmark for land use and land cover classification. IEEE Journal of Selected Topics in Applied Earth Observations and Remote Sensing 12(7):2217--2226

\bibitem[{Hinton et~al(2015)Hinton, Vinyals, and Dean}]{Knowledge_Distillation1}
Hinton G, Vinyals O, Dean J (2015) Distilling the knowledge in a neural network. arXiv preprint arXiv:150302531 2

\bibitem[{Hu et~al(2021)Hu, Tang, Miao, Hua, and Zhang}]{DistillIncre}
Hu X, Tang K, Miao C, Hua XS, Zhang H (2021) Distilling causal effect of data in class-incremental learning. In: CVPR, pp 3957--3966

\bibitem[{Jia et~al(2021)Jia, Yang, Xia, Chen, Parekh, Pham, Le, Sung, Li, and Duerig}]{jia2021scaling}
Jia C, Yang Y, Xia Y, Chen YT, Parekh Z, Pham H, Le Q, Sung YH, Li Z, Duerig T (2021) Scaling up visual and vision-language representation learning with noisy text supervision. In: ICML, pp 4904--4916

\bibitem[{Jia et~al(2022)Jia, Tang, Chen, Cardie, Belongie, Hariharan, and Lim}]{VPT}
Jia M, Tang L, Chen BC, Cardie C, Belongie S, Hariharan B, Lim SN (2022) Visual prompt tuning. In: ECCV

\bibitem[{Khattak et~al(2023)Khattak, Rasheed, Maaz, Khan, and Khan}]{MaPLe}
Khattak MU, Rasheed H, Maaz M, Khan S, Khan FS (2023) Maple: Multi-modal prompt learning. In: Proceedings of the IEEE/CVF Conference on Computer Vision and Pattern Recognition, pp 19113--19122

\bibitem[{Kim et~al(2021)Kim, Son, and Kim}]{Vilt}
Kim W, Son B, Kim I (2021) Vilt: Vision-and-language transformer without convolution or region supervision. In: International Conference on Machine Learning, PMLR, pp 5583--5594

\bibitem[{Krause et~al(2013)Krause, Stark, Deng, and Fei-Fei}]{StanfordCars}
Krause J, Stark M, Deng J, Fei-Fei L (2013) 3d object representations for fine-grained categorization. In: ICCV-W, pp 554--561

\bibitem[{Li et~al(2021{\natexlab{a}})Li, Weinberger, Belongie, Koltun, and Ranftl}]{li2021language}
Li B, Weinberger KQ, Belongie S, Koltun V, Ranftl R (2021{\natexlab{a}}) Language-driven semantic segmentation. In: ICLR

\bibitem[{Li and Wang(2023)}]{zeroi2v}
Li X, Wang L (2023) Zeroi2v: Zero-cost adaptation of pre-trained transformers from image to video. arXiv preprint arXiv:231001324

\bibitem[{Li and Liang(2021)}]{li2021prefix}
Li XL, Liang P (2021) Prefix-tuning: Optimizing continuous prompts for generation. In: ACL, pp 4582--4597

\bibitem[{Li et~al(2021{\natexlab{b}})Li, Liang, Zhao, Cui, Ouyang, Shao, Yu, and Yan}]{li2021supervision}
Li Y, Liang F, Zhao L, Cui Y, Ouyang W, Shao J, Yu F, Yan J (2021{\natexlab{b}}) Supervision exists everywhere: A data efficient contrastive language-image pre-training paradigm. In: ICLR

\bibitem[{Li and Hoiem(2016)}]{LearnWoForget}
Li Z, Hoiem D (2016) Learning without forgetting. In: ECCV, pp 614--629

\bibitem[{Liu et~al(2021)Liu, Ji, Fu, Du, Yang, and Tang}]{P-tuning}
Liu X, Ji K, Fu Y, Du Z, Yang Z, Tang J (2021) P-tuning v2: Prompt tuning can be comparable to fine-tuning universally across scales and tasks. arXiv preprint arXiv:211007602

\bibitem[{Liu et~al(2020)Liu, Su, Liu, Schiele, and Sun}]{MCIncre}
Liu Y, Su Y, Liu AA, Schiele B, Sun Q (2020) Mnemonics training: Multi-class incremental learning without forgetting. In: CVPR, pp 12245--12254

\bibitem[{Lu et~al(2019)Lu, Batra, Parikh, and Lee}]{Vilbert}
Lu J, Batra D, Parikh D, Lee S (2019) Vilbert: Pretraining task-agnostic visiolinguistic representations for vision-and-language tasks. Advances in neural information processing systems 32

\bibitem[{Lu et~al(2022)Lu, Liu, Zhang, Liu, and Tian}]{lu2022prompt}
Lu Y, Liu J, Zhang Y, Liu Y, Tian X (2022) Prompt distribution learning. In: CVPR, pp 5206--5215

\bibitem[{Maji et~al(2013)Maji, Rahtu, Kannala, Blaschko, and Vedaldi}]{FGVCAircraft}
Maji S, Rahtu E, Kannala J, Blaschko M, Vedaldi A (2013) Fine-grained visual classification of aircraft. arXiv preprint arXiv:13065151

\bibitem[{Mokady et~al(2021)Mokady, Hertz, and Bermano}]{mokady2021clipcap}
Mokady R, Hertz A, Bermano AH (2021) Clipcap: Clip prefix for image captioning. arXiv preprint arXiv:211109734

\bibitem[{Nilsback and Zisserman(2008)}]{OxfordFlowers}
Nilsback ME, Zisserman A (2008) Automated flower classification over a large number of classes. In: ICVGIP, pp 722--729

\bibitem[{Parkhi et~al(2012)Parkhi, Vedaldi, Zisserman, and Jawahar}]{OxfordPets}
Parkhi OM, Vedaldi A, Zisserman A, Jawahar C (2012) Cats and dogs. In: CVPR, pp 3498--3505

\bibitem[{Phuong and Lampert(2019)}]{Knowledge_Distillation2}
Phuong M, Lampert C (2019) Towards understanding knowledge distillation. In: ICML, pp 5142--5151

\bibitem[{Qin and Joty(2022)}]{regularRL}
Qin C, Joty S (2022) Continual few-shot relation learning via embedding space regularization and data augmentation. In: ACL, pp 2776--2789

\bibitem[{Radford et~al(2021)Radford, Kim, Hallacy, Ramesh, Goh, Agarwal, Sastry, Askell, Mishkin, Clark et~al}]{CLIP}
Radford A, Kim JW, Hallacy C, Ramesh A, Goh G, Agarwal S, Sastry G, Askell A, Mishkin P, Clark J, et~al (2021) Learning transferable visual models from natural language supervision. In: ICML, pp 8748--8763

\bibitem[{Rebuffi et~al(2017)Rebuffi, Kolesnikov, Sperl, and Lampert}]{icarl}
Rebuffi SA, Kolesnikov A, Sperl G, Lampert CH (2017) icarl: Incremental classifier and representation learning. In: CVPR, pp 2001--2010

\bibitem[{Ren et~al(2022)Ren, Li, Ren, Zhao, and Sun}]{OpennessOfCLIP}
Ren S, Li L, Ren X, Zhao G, Sun X (2022) Rethinking the openness of clip. arXiv preprint arXiv:220601986

\bibitem[{Riemer et~al(2019)Riemer, Cases, Ajemian, Liu, Rish, Tu, and Tesauro}]{MTMI}
Riemer M, Cases I, Ajemian R, Liu M, Rish I, Tu Y, Tesauro G (2019) Learning to learn without forgetting by maximizing transfer and minimizing interference. In: ICLR

\bibitem[{Shen et~al(2021)Shen, Li, Tan, Bansal, Rohrbach, Chang, Yao, and Keutzer}]{shen2021much}
Shen S, Li LH, Tan H, Bansal M, Rohrbach A, Chang KW, Yao Z, Keutzer K (2021) How much can clip benefit vision-and-language tasks? In: ICLR

\bibitem[{Shi et~al(2022)Shi, Hayat, Wu, and Cai}]{shi2022proposalclip}
Shi H, Hayat M, Wu Y, Cai J (2022) Proposalclip: Unsupervised open-category object proposal generation via exploiting clip cues. In: CVPR, pp 9611--9620

\bibitem[{Shu et~al(2022)Shu, Nie, Huang, Yu, Goldstein, Anandkumar, and Xiao}]{shu2022test}
Shu M, Nie W, Huang DA, Yu Z, Goldstein T, Anandkumar A, Xiao C (2022) Test-time prompt tuning for zero-shot generalization in vision-language models. arXiv preprint arXiv:220907511

\bibitem[{Soomro et~al(2012)Soomro, Zamir, and Shah}]{UCF101}
Soomro K, Zamir AR, Shah M (2012) Ucf101: A dataset of 101 human actions classes from videos in the wild. arXiv preprint arXiv:12120402

\bibitem[{Su et~al(2019)Su, Zhu, Cao, Li, Lu, Wei, and Dai}]{Vl-bert}
Su W, Zhu X, Cao Y, Li B, Lu L, Wei F, Dai J (2019) Vl-bert: Pre-training of generic visual-linguistic representations. arXiv preprint arXiv:190808530

\bibitem[{Sun et~al(2023)Sun, Fang, Wu, Wang, and Cao}]{EVA-CLIP}
Sun Q, Fang Y, Wu L, Wang X, Cao Y (2023) Eva-clip: Improved training techniques for clip at scale. arXiv preprint arXiv:230315389

\bibitem[{Tan and Bansal(2019)}]{Lxmert}
Tan H, Bansal M (2019) Lxmert: Learning cross-modality encoder representations from transformers. arXiv preprint arXiv:190807490

\bibitem[{Tang et~al(2021)Tang, Wang, Liu, Rao, Li, and Li}]{tang2021clip4caption}
Tang M, Wang Z, Liu Z, Rao F, Li D, Li X (2021) Clip4caption: Clip for video caption. In: ACM International Conference on Multimedia, pp 4858--4862

\bibitem[{Vaswani et~al(2017)Vaswani, Shazeer, Parmar, Uszkoreit, Jones, Gomez, Kaiser, and Polosukhin}]{Attention}
Vaswani A, Shazeer N, Parmar N, Uszkoreit J, Jones L, Gomez AN, Kaiser {\L}, Polosukhin I (2017) Attention is all you need. Advances in neural information processing systems 30:5998--6008

\bibitem[{Wang et~al(2022{\natexlab{a}})Wang, Huang, and Hong}]{wang2022s}
Wang Y, Huang Z, Hong X (2022{\natexlab{a}}) S-prompts learning with pre-trained transformers: An occam's razor for domain incremental learning. arXiv preprint arXiv:220712819

\bibitem[{Wang et~al(2022{\natexlab{b}})Wang, Lu, Li, Tao, Guo, Gong, and Liu}]{wang2022cris}
Wang Z, Lu Y, Li Q, Tao X, Guo Y, Gong M, Liu T (2022{\natexlab{b}}) Cris: Clip-driven referring image segmentation. In: CVPR, pp 11686--11695

\bibitem[{Wang et~al(2022{\natexlab{c}})Wang, Zhang, Lee, Zhang, Sun, Ren, Su, Perot, Dy, and Pfister}]{wang2022learning}
Wang Z, Zhang Z, Lee CY, Zhang H, Sun R, Ren X, Su G, Perot V, Dy J, Pfister T (2022{\natexlab{c}}) Learning to prompt for continual learning. In: CVPR, pp 139--149

\bibitem[{Xiao et~al(2010)Xiao, Hays, Ehinger, Oliva, and Torralba}]{SUN397}
Xiao J, Hays J, Ehinger KA, Oliva A, Torralba A (2010) Sun database: Large-scale scene recognition from abbey to zoo. In: CVPR, pp 3485--3492

\bibitem[{Xu et~al(2023)Xu, Zhu, Zhang, Shen, Liao, Chen, Wu, and Wang}]{xu2023dpl}
Xu C, Zhu Y, Zhang G, Shen H, Liao Y, Chen X, Wu G, Wang L (2023) Dpl: Decoupled prompt learning for vision-language models. arXiv preprint arXiv:230810061

\bibitem[{Yuan et~al(2021)Yuan, Chen, Chen, Codella, Dai, Gao, Hu, Huang, Li, Li et~al}]{yuan2021florence}
Yuan L, Chen D, Chen YL, Codella N, Dai X, Gao J, Hu H, Huang X, Li B, Li C, et~al (2021) Florence: A new foundation model for computer vision. arXiv preprint arXiv:211111432

\bibitem[{Zhai et~al(2022)Zhai, Wang, Mustafa, Steiner, Keysers, Kolesnikov, and Beyer}]{zhai2022lit}
Zhai X, Wang X, Mustafa B, Steiner A, Keysers D, Kolesnikov A, Beyer L (2022) Lit: Zero-shot transfer with locked-image text tuning. In: CVPR, pp 18123--18133

\bibitem[{Zhang et~al(2023)Zhang, Zhu, Wang, Chen, Wu, and Wang}]{ema}
Zhang G, Zhu Y, Wang H, Chen Y, Wu G, Wang L (2023) Extracting motion and appearance via inter-frame attention for efficient video frame interpolation. In: Proceedings of the IEEE/CVF Conference on Computer Vision and Pattern Recognition, pp 5682--5692

\bibitem[{Zhang et~al(2021)Zhang, Fang, Gao, Zhang, Li, Dai, Qiao, and Li}]{zhang2021tip}
Zhang R, Fang R, Gao P, Zhang W, Li K, Dai J, Qiao Y, Li H (2021) Tip-adapter: Training-free clip-adapter for better vision-language modeling. arXiv preprint arXiv:211103930

\bibitem[{Zhou et~al(2022{\natexlab{a}})Zhou, Yang, Loy, and Liu}]{CoCoOp}
Zhou K, Yang J, Loy CC, Liu Z (2022{\natexlab{a}}) Conditional prompt learning for vision-language models. In: CVPR, pp 16816--16825

\bibitem[{Zhou et~al(2022{\natexlab{b}})Zhou, Yang, Loy, and Liu}]{CoOp}
Zhou K, Yang J, Loy CC, Liu Z (2022{\natexlab{b}}) Learning to prompt for vision-language models. International Journal of Computer Vision 130(9):2337--2348

\bibitem[{Zhu et~al(2022)Zhu, Niu, Han, Wu, and Zhang}]{ProGrad}
Zhu B, Niu Y, Han Y, Wu Y, Zhang H (2022) Prompt-aligned gradient for prompt tuning. arXiv preprint arXiv:220514865

\bibitem[{Zhu et~al(2024)Zhu, Zhang, Tan, Wu, and Wang}]{dualdetr}
Zhu Y, Zhang G, Tan J, Wu G, Wang L (2024) Dual detrs for multi-label temporal action detection. In: Proceedings of the IEEE/CVF Conference on Computer Vision and Pattern Recognition, pp 18559--18569

\end{thebibliography}

\end{document}